\documentclass[sigconf]{aamas}  

\AtBeginDocument{%
  \providecommand\BibTeX{{%
    \normalfont B\kern-0.5em{\scshape i\kern-0.25em b}\kern-0.8em\TeX}}}
    
\usepackage{booktabs}    
\usepackage{flushend} 

\def\fighome{figures}

\usepackage{url}
\usepackage{dsfont}
\usepackage{xspace}
\usepackage{hyperref}
\hypersetup{
    colorlinks=true,      
    urlcolor=blue
}
\usepackage[outdir=./]{epstopdf}
\usepackage{graphicx,float,pgfplots,wrapfig,sidecap,lipsum}
\usepackage{tabularx}
\usepackage{booktabs}
\usepackage{paralist}
\usepackage[linesnumbered,algoruled,boxed,lined,noend]{algorithm2e}
\usepackage{amsfonts}
\usepackage{amsthm}
\usepackage{amsmath}
\usepackage{amssymb}
\usepackage{enumitem}
\usepackage{xcolor}
\usepackage[font=normal,labelfont=bf]{caption}
\usepackage{mathtools}

\usepackage{bm}

\usepackage{tablefootnote}
\usepackage{subcaption}
\usepackage{tikz}
\usetikzlibrary{fit}
\usetikzlibrary{calc,shapes}
\usetikzlibrary{decorations.pathmorphing} 
\usetikzlibrary{fit}					
\usetikzlibrary{backgrounds}	

\usepackage{bm}
\usepackage{mathrsfs} 
\usepackage{stmaryrd}
\usepackage[utf8]{inputenc}
\usepackage[english]{babel}
\usepackage{diagbox}

\usepackage[utf8]{inputenc}
\usepackage{pgfplots}
\pgfplotsset{compat=newest}
\usepgfplotslibrary{groupplots}
\usepgfplotslibrary{dateplot}

\newcommand*{\mymatrix}[1]{\bm{#1}}
\newcommand*{\myvector}[1]{\bm{#1}}

\newcommand{\ourmod}{{GIM}\xspace}
\newcommand{\ourmodfull}{{Greedy Inference Model}\xspace}
\newcommand{\dmname}{{dynamic matrix}\xspace}
\newcommand{\dmnames}{{dynamic matrices}\xspace}
\newcommand{\avgrew}{{AvgReward}\xspace}
\newcommand{\pavgrew}{{PostAvgReward}\xspace}
\newcommand{\numepi}{{TotalEps}\xspace}
\newcommand{\walkname}{{$\beta$-curious walking}\xspace}

\newcommand{\states}{\mathcal{S}}
\newcommand{\actions}{\mathcal{A}}
\newcommand{\dmat}{\mymatrix{M}}
\newcommand{\dmats}{M}
\newcommand{\admat}{\mymatrix{\hat{M}}}
\newcommand{\rdmat}{\mymatrix{\widetilde{M}}}
\newcommand{\rdmats}{\widetilde{M}}
\newcommand{\dmatset}{\{\mymatrix{M}^k\}_{k=1}^{S+1}}
\newcommand{\admatset}{\{\mymatrix{\hat{M}}^k\}_{k=1}^{S+1}}

\newcommand{\krange}{1,2,\cdots,S+1}

\newcommand{\mask}{\mymatrix{P}^\mathcal{K}}
\newcommand{\masks}{P^\mathcal{K}}
\newcommand{\kset}{\mathcal{K}}
\newcommand{\nkset}{\bar{\mathcal{K}}}

\newcommand{\noisem}{\mymatrix{Z}}

\newcommand{\fraction}{\rho}

\newcommand{\minsa}{M_{in}}
\newcommand{\maxsa}{M_{ax}}

\DeclareMathOperator*{\argmax}{arg\,max}

\SetKwInput{Input}{Input}
\SetKwInput{Output}{Output}
\SetKwInput{Parameter}{Parameters}
\SetKwFunction{Ma}{Mature}
\SetKwFunction{Cp}{Complete}
\SetKwFunction{Up}{Update}
\SetKwFunction{MC}{MatComp}
\SetKwProg{Fn}{Function}{:}{}

\def\beq{\begin{equation}}

\def\eeq{\end{equation}\noindent}

\newcommand{\bp}{\begin{psfrags}}
\newcommand{\ep}{\end{psfrags}}
\newcommand{\bc}{\begin{center}}
\newcommand{\ec}{\end{center}}

\newcommand*{\tran}{^{\mkern-1.5mu\mathsf{T}}}

\newtheorem{condition}{Condition}

\newtheorem{assumption}{Assumption}

\floatname{algorithm}{Procedure}

\newcommand{\mytitle}{Can Agents Learn by Analogy?
An Inferable  Model for PAC Reinforcement Learning}

\setcopyright{ifaamas}  
\copyrightyear{2020} 
\acmYear{2020} 
\acmDOI{} 
\acmPrice{} 
\acmISBN{} 
\acmConference[AAMAS'20]{Proc.\@ of the 19th International Conference on Autonomous Agents and Multiagent Systems (AAMAS 2020)}{May 9--13, 2020}{Auckland, New Zealand}{B.~An, N.~Yorke-Smith, A.~El~Fallah~Seghrouchni, G.~Sukthankar (eds.)}  

\begin{document}

\title{\mytitle}  


\author{Yanchao Sun}
\affiliation{%
  \institution{University of Maryland}
  \city{College Park} 
  \state{MD} 
  \postcode{20740}
}
\email{ycs@cs.umd.edu}
\author{Furong Huang}
\affiliation{%
  \institution{University of Maryland}
  \city{College Park} 
  \state{MD} 
  \postcode{20740}
}
\email{furongh@cs.umd.edu}

\begin{abstract}
Model-based reinforcement learning algorithms make decisions by building and utilizing a model of the environment. 
However, none of the existing algorithms attempts to infer the dynamics of any state-action pair from known state-action pairs before meeting it for sufficient times.
We propose a new model-based method called \ourmodfull (\ourmod) that infers the unknown dynamics from known dynamics based on the internal spectral properties of the environment. 
In other words, \ourmod can ``learn by analogy''. 
We further introduce a new exploration strategy which ensures that the agent rapidly and evenly visits unknown state-action pairs.
\ourmod is much more computationally efficient than state-of-the-art model-based algorithms, as the number of dynamic programming operations is independent of the environment size. 
Lower sample complexity could also be achieved under mild conditions compared against methods without inferring.
Experimental results demonstrate the effectiveness and efficiency of \ourmod in a variety of real-world tasks.
\end{abstract}

%

\keywords{Model-based reinforcement learning; Spectral method; Sample complexity; Computational complexity} 

\maketitle


\section{Introduction}
\label{sec:intro}

In Reinforcement Learning (RL)\cite{sutton2018reinforcement}, an agent interacts with the environment by taking actions and receiving rewards or payoffs to figure out a policy that maximizes the total rewards. 
Recently, RL has been successfully applied in many fields such as robotics \cite{kober2013reinforcement}, games \cite{mnih2015human}, recommendation systems \cite{zhang2019hierarchical}, etc. 
However, the high sample complexity and cost of computational resources prevent RL algorithms from being successfully deployed in many real-world tasks.

We call the RL algorithms which explicitly learn a model from experiences \textit{model-based}, and algorithms that directly learn from interactions without any model \textit{model-free}. Although these two types of algorithms are both effective in learning, they usually differ in terms of \textit{sample complexity}, \textit{computational complexity} and \textit{space complexity}, which respectively measure the amount of interactions, computations, and memory an algorithm needs in RL tasks. Model-based algorithms
are more sample efficient, but cost more computations and space. In contrast, model-free algorithms
save computations and space, but usually need more samples/experience to learn, and easily get trapped in local optima.



In this paper, we focus on model-based algorithms due to the following two reasons. First, model-based algorithms make more efficient use of samples than model-free ones. Most existing PAC-MDP algorithms\footnote{An algorithm is PAC-MDP (Probably Approximately Correct in Markov Decision Processes) if its sample complexity is polynomial in the environment size and approximation parameters with high probability.} are model-based. 
Second, the learned model is an abstraction of the environment, and can be easily transferred to other similar tasks \cite{brunskill2013sample}. 
For instance, if we change the reward of a state, only the reward value of the state should be changed in the learned model in model-based methods. However, for model-free methods, many state and action values will be affected.




Our goal is to find a method that can reduce both the sample and computational complexity of model-based methods. We focus on the following challenges:
\begin{itemize}[itemsep=0in]
	\item \emph{High stochasticity.} The transitions among states are usually stochastic. The highly stochastic transitions require a large number of trials and errors to reach the right decisions. Can we avoid visiting the highly stochastic transitions and still achieve a good policy?
	\item \emph{Dilemma between sample and computational complexity.} Utilizing samples in an efficient manner requires more operations, while pursuing high speed may sacrifice accuracy and lead to more errors.  Can we achieve both sample and computational efficiency?
	\item \emph{Interplay of exploration and exploitation.} The trade-off between exploration and exploitation is a crucial problem in RL. Should we always iteratively alternate between exploration and exploitation as implemented by most existing algorithms?
\end{itemize}

Recent efforts \cite{szita2010model,strehl2012incremental,jiang2018pac}  improve the sample and computational efficiency of model-based RL algorithms from various perspectives. 
However, the internal structure of the underlying Markov Decision Process (MDP) is ignored. 
The internal structure of MDPs refers to the spectral properties of the transition and reward distributions. 
More explicitly, we observed that many MDPs have locally or globally interrelated dynamics, resulting from the existence of similar states, actions or transitions. 
For example, consider a simple $2 \times 3$ grid world in Table~\ref{tbl:example}(left) where the agent can go up, down, left and right. 
The floor is slippery so when it wants to go up, it may slip to either left or right with probability 0.2. 
(Slipping also happens for other actions).
If there is a wall in the objective direction, it stays in the current state. 
Table~\ref{tbl:example}(right) is the transition table from other states to state 2; the entries are the transition probabilities from state-action pairs to state 2. 
The rows of state 4 and 6 are omitted because state 2 is not immediately reachable from state 4 or 6. 
We find that the rows of state 2 and 5 are exactly the same, and the rank of this matrix is 3, smaller than the number of states or actions. 
This phenomenon becomes more ubiquitous in larger environments.

\begin{table}[!htb]
    \begin{minipage}{.3\linewidth}
      \centering
        \begin{tabular}{|c|c|c|}
		\hline
		1 & 2 & 3\\ \hline
		4 & 5 & 6 \\ \hline
		\end{tabular}
    \end{minipage}%
    \begin{minipage}{.7\linewidth}
      \centering
        \begin{tabular}{c|cccc}
            & up & down & left & right \\ \hline
            1 & 0.2 & 0.2 & 0 & 0.6 \\ 
            2 & 0.6 & 0 & 0.2 & 0.2 \\ 
            3 & 0.2 & 0.2 & 0.6 & 0 \\ 
            5 & 0.6 & 0 & 0.2 & 0.2
        \end{tabular}
    \end{minipage} 
    \caption{A grid world example and the transition table to state 2.}
    \label{tbl:example}
    \vspace{-2em}
\end{table}

Due to the existence of such similar structures, we do not have to learn every state and action thoroughly, and the exploration can be much more efficient. 
We propose to explore a subset of ``essential'' transition dynamics, then infer the remaining dynamics using spectral methods, to achieve sample and computational efficiency.

In this paper, we propose a novel model-based RL algorithm called \ourmodfull (\ourmod), which utilizes the structural and spectral characteristics of MDPs and thus expedites the learning process. 
\ourmod introduces a novel exploration strategy to discover the unknowns efficiently, and a spectral method to estimate the entire model with the knowledge of a fraction of the model. 
The core idea of \ourmod can be applied to any model-based methods. 
We prove that \ourmod is PAC-MDP, and it has significantly lower computational complexity and potentially lower sample complexity than state-of-the-art model-based algorithms.
Systematic empirical studies demonstrate that \ourmod outperforms both model-based and model-free state-of-the-art approaches on a wide variety of tasks.

Our contributions are summarized as follows:
\begin{itemize}
	\item  To the best of our knowledge, we are \textbf{the first to estimate the model by utilizing the internal structure of the MDPs with guaranteed accuracy}. We avoid directly estimating the highly stochastic transitions, which is sample-consuming. 
	\item We show that \ourmod can significantly \textbf{reduce the computational cost}, as the number of dynamic programming operations is independent of the environment size. We also prove \ourmod could \textbf{improve the sample efficiency} of model-based algorithms.
	\item We propose a \textbf{new mechanism to address the exploration and exploitation dilemma}. By using a new exploration strategy (\walkname), \ourmod takes fewer exploration steps in total than existing methods.
\end{itemize}

\section{Related Work}
\label{sec:related}

\subsection{RL Algorithms}

\noindent \textbf{Model-based algorithms.} 
Model-based algorithms like $E^3$~\cite{kearns2002near}, RMax~\cite{brafman2003rmax} and MBIE~\cite{strehl2005a} construct a model from interactions, storing the transition probabilities and rewards of every state and action pair, and then make predictions with the model. 
Followup works improve the efficiency of aforementioned model-based algorithms. 
RTDP-RMAX and RTDP-MBIE~\cite{strehl2012incremental} reduce the number of updates and achieve lower computational complexity, with minor performance loss on the accumulated rewards achieved. 
MORMAX~\cite{szita2010model} modifies RMax algorithm and reduces the sample complexity~\cite{kakade2003sample,dann2015sample} by maintaining an imperfect model, but the model estimation is not accurate which prevents accurate rewards predictions for some state-action pairs. 
\cite{jiang2018pac} proposes a method with no dependence on the size of the state-action space, but it assumes that an approximate imperfect model is given. 

\noindent \textbf{Model-free algorithms.} Model-free algorithms \cite{watkins1992q,NIPS2010_3964,williams1992simple} decide what actions to take based on the trajectory/history. Delayed Q-learning~\cite{strehl2006pac} is a special model-free algorithm as it is PAC-MDP, whose sample complexity depends linearly on the state and the action number. However Delayed Q-learning has higher dependence on the discount factor $\gamma$ and the error tolerance $\epsilon$ than RMax.

\noindent \textbf{Deep RL algorithms.}
Recently, researchers have made significant progress by combining deep learning with both model-based or model-free RL~\cite{mnih2015human,van2016deep} and achieving impressive empirical performance. 
However theoretical understanding of deep learning , and thus deep RL, remains unsettled. 
Deep RL, usually applied for large-scale decision-making problems, requires large number of training examples, which are not practical for tasks with limited training examples.

\noindent \textbf{PAC RL.}
A key goal of RL algorithms is to maximize the reward with as few samples as possible. The sample-efficiency of RL algorithms can be measured by the PAC performance metric (sample complexity) first formally defined in~\cite{kakade2003sample}. \cite{dann2015sample} derives a tighter PAC upper bound for episodic fixed-horizon RL tasks. Recently, more strict metrics like Uniform-PAC and IPOC~\cite{dann2017unifying,dann2018policy} are proposed to measure the performance of RL algorithms. And by computing certificates for optimistic RL algorithms~\cite{dann2018policy}, minimax-optimal PAC bounds up to lower-order terms are achieved under certain conditions.



\vspace*{-1em}
\subsection{Spectral Methods}
\noindent \textbf{Matrix completion.} The spectral method we will use in this paper is mainly the well-studied matrix completion. It is proved that we can recover a matrix with only a fraction of its (noisy) entries~\cite{keshavan2010matrix,candes2010matrix}. 

\noindent \textbf{Spectral methods and RL.} Spectral methods have been applied in RL in the learning of POMDP (Partially Observable Markov Decision Process)~\cite{azizzadenesheli2016reinforcement} and ROMDP (Rich-Observation Markov Decision Process)~\cite{azizzadenesheli2016romdp}, where a multi-view model~\cite{anandkumar2014tensor} is used. Researchers discover that knowledge can be transferred between tasks, domains or agents~\cite{ferguson2006proto,lazaric2013sequential,bromuri2012tensor} using spectral methods. Moreover, some recent works propose new learning algorithms by constructing certain low-rank models \cite{ong2015value,jiang2017contextual,boots2011closing}, where spectral methods are involved. 

\noindent \textbf{Low-rank transition model.} There is a line of works learning the low-rank structure of the transition models~\cite{yang2019reinforcement,duan2018adaptive,yang2019sample,li2018estimation,jin2019provably}, although we focus on different low-rank objects and use different models as well as assumptions.


\section{Notations and Problem Setup}
\label{sec:prelim}

\subsection{Notations for RL}
In this paper, we focus on episodic, discrete-time, and fixed horizon MDPs with finite state and action spaces. 
A Markov decision process (MDP) is defined as a tuple $<\states,\actions,p(\cdot|\cdot,\cdot),r(\cdot, \cdot),\mu>$, where $\mathcal{S}$ is the state space (with cardinality $S$); $\mathcal{A}$ is the action space (with cardinality $A$); $p(\cdot|\cdot,\cdot)$ is the transition probability function with $p(s_k | s_i, a_j)$ representing the probability of transiting to state $s_k$ from state $s_i$ by taking action $a_j$; $r(\cdot, \cdot)$ is the reward function with $r(s_i, a_j)$ recording the reward one can get by taking action $a_j$ in state $s_i$; $\mu$ is the initial state distribution. $p(\cdot|\cdot,\cdot)$ and $r(\cdot, \cdot)$ together are called the dynamics of the MDP. 
We use $H$ to denote the horizon (number of steps one can take in an episode) of an MDP.

\begin{definition}[Dynamic Matrices]
	Given an MDP $M$ denoted by tuple $<\states, \actions, p(\cdot|\cdot,\cdot), r(\cdot, \cdot), \mu>$, we define $S+1$ dynamic matrices $\{\dmat^{s}\}_{s\in\states}$ and $\dmat^r$. $\{\dmat^{s}\}_{s\in\states}$ are called transition \dmnames, where $\dmat^{s}_{ij}=p(s|s_i,a_j)$ for all $s \in \states$. 
 	$\dmat^r$ is called reward dynamic matrix in which $\dmat^r_{ij} = r(s_i, a_j)$.
\end{definition}


The empirical estimations of the dynamic matrices are:
\begin{equation}\label{eq:admat}	
	 \admat^{s}_{ij} = \frac{n(s | s_i, a_j)}{n(s_i,a_j)}\ \  \forall s  \text{ and }
	 \admat^{r}_{ij} = \frac{R(s_i,a_j)}{n(s_i,a_j)},
\end{equation}
where $n(s_i,a_j)$ is the total number of visits to state-action pair $(s_i,a_j)$, $n(s | s_i, a_j)$ the total number of transitions from $s_i$ to $s$ by taking action $a_j$ and $R(s_i,a_j)$ the total rewards \footnote{$R(s_i,a_j)$ is the empirical total rewards gained by visiting $(s_i,a_j)$ in the history, and is different from $r(s_i,a_j)$.} for $(s_i,a_j)$.
The empirical dynamic matrix $\admat$ is an approximation of the corresponding dynamic matrix $\dmat$, so we have $\admat = \dmat + \noisem$ where $\noisem$ is a noise matrix. The more observations we have, the more accurate the approximation is. 

Our main goal is to recover every $\dmat$ based on $\admat$. More explicitly, given the empirical dynamic matrix $\admat$, the algorithm should return a matrix $\rdmat$ that is $\epsilon$-close to the original $\dmat$, i.e., $\| \rdmat - \dmat \| \leq \epsilon$. 

The value function of a policy $\pi$ for a given MDP $M$ with horizon $H$ is the expected average reward
$V^\pi_M = \mathbb{E}_{s_0 \sim \mu}[\frac{1}{H}\sum_{h=0}^{H-1} r(s_h, \pi(s_h))].$
The optimal policy $\pi^*$ is the policy that achieves the largest possible value $V^*_M$.
In an RL task, an agent searches for the optimal policy by interacting with the MDP.  
The general goal of RL algorithms is to learn the optimal policy for any given MDP with as few interactions as possible. A widely-used framework to evaluate the performance of RL algorithms is \textit{sample complexity of exploration} \cite{kakade2003sample}, or \textit{sample complexity} for short. 

\begin{definition}[Sample complexity of exploration]
\label{def:sc}
For any $\epsilon>0$ and $0 < \delta < 1$, and at any episode $t$, 
if the policy $\pi_t$ generated by an RL algorithm $L$ satisfies $V^*-V^{\pi_t} \leq \epsilon$, we say $L$ is near-optimal at episode $t$. If with probability at least $1-\delta$, the total number of episodes that $L$ is not near-optimal is upper bounded by a function $\zeta(\epsilon, \delta)$, then $\zeta$ is called the sample complexity of $L$.
\end{definition}

Intuitively, sample complexity illustrates the number of steps in which the agent does not act near-optimally. 




\subsection{Notations for Spectral Methods}
Incoherence~\cite{keshavan2010matrix} of a matrix is an important property that demonstrates the ``sparsity'' of the singular vectors of the matrix: all coordinates of each singular vector are of comparable magnitude (a.k.a., a dense singular vector) vs just a few coordinates having significantly larger magnitudes (a.k.a., a sparse singular vector).

\begin{definition}[($\mu_0$,$\mu_1$)-incoherence]
\label{def:incohe}
	 A matrix $\mymatrix{M} \in \mathbb{R}^{m \times n}$ with rank $r$ has SVD $\mymatrix{M} = \mymatrix{U} \mymatrix{\Sigma} \mymatrix{V\tran}$, where $\mymatrix{U}$ and $\mymatrix{V}$ are orthonormal. We say $\mymatrix{M}$ is $(\mu_0,\mu_1)$-incoherent if (1) for all $i \in [m]$, $j \in [n]$ we have $\sum_{k=1}^r U_{ik}^2 \leq \mu_0 r$, and $\sum_{k=1}^r V_{jk}^2 \leq \mu_0 r$; (2) There exist $\mu_1$ such that $|\sum_{k=1}^r U_{ik} (\Sigma_k / \Sigma_1) V_{jk}| \leq \mu_1 \sqrt{r}$, where $\Sigma_k$ is the $k$-th singular value of $\mymatrix{M}$.
\end{definition}

The smaller $\mu_0$ and $\mu_1$ are, the more spread-out the singular vectors are. As a result, matrix completion methods require a smaller number of known entries to confidently recover the entire matrix.
See Appendix\footnote{The Appendix of this paper is in \url{https://arxiv.org/abs/1912.10329}}~\ref{app:mc} for details about matrix completion and incoherence.


\section{Motivations}
\label{sec:motivations}
Before the formal introduction of our proposed learning algorithm, we consider two questions: 

\noindent \textbf{Is it necessary to learn every state-action pair from scratch?}\\
The key to RL is to evaluate the value of every state and action, with or without a model. 
The agent makes observations of each state-action pair, accumulates experience, and estimates the model or the values. It knows nothing about a state-action pair before meeting it. However, is it necessary to learn every new state-action pair from scratch? As humans, we can learn by analogy. 
For example, if one has jumped out of a window on the second floor and got injured, he will learn never to jump from another window on the third floor, because he accumulates knowledge from his previous experience and finds the internal connections between these two situations.
But existing RL agents, which are not able to analyze new states, tend to make the same mistakes in similar situations.

Therefore, in this work, we extract and use the internal connections of the environment via spectral methods to reduce the unnecessary trials and errors for the agent.


\noindent \textbf{Should we always interleave exploration and exploitation?}\\
The exploration-exploitation dilemma has been intensively studied for decades and remains unsolved. 
In RL, exploration is to try the unknowns, while exploitation maximizes rewards based on the current knowledge. 
Most RL algorithms interleave (or alternate between) exploration and exploitation. An example is the widely-used $\varepsilon$-greedy exploration method, which chooses actions greedily with respect to the action values with probability $1-\varepsilon$ (exploit), and randomly chooses actions with probability $\varepsilon$ (explore). Moreover, many model-based algorithms, such as $E^3$ and RMax, choose actions with the maximum value in known states, and execute the action that has been tried the fewest times in unknown states.
However, is this interleaving the only manner to get the optimal results? 
If the agent ``greedily'' chooses the most rewarding action when it knows little about the whole environment, it usually misses the largest possible reward in the long run. 
Can the agent ignore the short-term benefits in the beginning, and keep exploring before it gains enough knowledge? 

In this work, we implement a two-phase algorithm in which exploitation follows after exploration, instead of interleaving the two. 
And we prove that our new method requires fewer samples and computations.
\section{Greedy Inference Model}
\label{sec:model}

In this section, we present a novel model-based RL algorithm called \ourmodfull (\ourmod) that considers the structural properties of MDPs and separates exploration from exploitation. More explicitly, two main ideas of \ourmod are (1) using matrix completion to recover/complete the \dmnames, and (2) greedily exploring the unknowns.

\subsection{Complete Unknowns with Knowns}
As in many model-based algorithms \cite{kearns2002near,brafman2003rmax}, we distinguish all state-action pairs as ``$m$-known'' or ``$m$-unknown'' (we will say known and unknown for short) pairs: a state-action pair is known if we have visited it for over $m$ times, so that the estimations for its transition probabilities and average reward are guaranteed to be nearly accurate with high probability. 
We use $\kset$ to denote the set of all known state-action pairs, and $\nkset$ for unknowns.


\begin{definition}[Known-ness Mask]
\label{def:mask}
	For an MDP $M$ with $S$ states and $A$ actions, the known-ness mask $\masks \in \mathbb{R}^{S \times A}$ is a binary matrix defined as
	\begin{equation}
		\masks_{i,j} = 
		\left\{
		\begin{aligned}
		& 1 & & \text{if }(s_i,a_j) \text{ is $m$-known} \\
		& 0 & & \text{otherwise}
		\end{aligned}
		\right.
	\end{equation}
\end{definition}

\noindent \textbf{Remark.} The summation of all entries of the known-ness mask is the total number of known state-action pairs in the MDP. Row sums and column sums are the numbers of known state-action pairs related to every state or action respectively.


As discussed in Motivations, unlike previous model-based methods such as RMax, we avoid the necessity of observing and gaining knowledge on every single state-action pair as the MDPs usually have some internal structure/pattern. 
We use matrix completion, a widely used spectral method,  to estimate the missing values in partially observed matrices. 
We now introduce how to complete the ``unknowns'' with ``knowns'' in an MDP. 

\subsubsection{Estimate Unknowns via Matrix Completion}
\label{sec:completion}

Matrix completion is the problem of recovering unknown entries in the matrix from a small fraction of its known (noisy) entries, which are randomly sampled from the matrix. 
When the matrix satisfies some assumptions that we will discuss later, the recovery is guaranteed to be accurate, even under noisy known entries. 
Based on matrix completion theory, \ourmod needs only a fraction of state-action pairs to be known to recover the unknown state-action pairs.
Now we formally define the matrix completion problem as the optimization problem: for every \dmname $\dmat$
\begin{align}
\label{eq:mc_problem}
\begin{array} { l l } 
	\text{minimize}_{\rdmat} &\| \mask \odot (\rdmat - \admat) \| \\
	\text{subject to } &rank(\rdmat) \leq r
\end{array}
\end{align}
where $\mask$ is the known-ness mask defined in Definition~\ref{def:mask}, $\odot$ denotes element-wise product, and $r$ is the rank of $\dmat$ or the upper bound of the rank. 

\subsubsection{Requirements for Accurate Completion}

Matrix completion makes it possible to know all the dynamics from some known state-action pairs, but the accuracy of completed dynamics
is determined by the structure of the matrix, as well as the number and the locations of known entries. 
In general, matrix completion with noisy observations requires (1) the number of known entries be greater than some threshold, and (2) the known entries be spread out randomly. 
We propose the following exploration strategy that conforms to the two requirements above.


\subsection{Greedily Explore the Environment}

To satisfy the two requirements on the known state-action pairs and guarantee the success of matrix completion, we propose a new exploration strategy called \textit{\walkname}. 

Let $\fraction$ denote the fraction of known state-action pairs over all state-action pairs. Therefore $\fraction SA = |\kset|$. We introduce the concept of $\fraction$-known state below.

\begin{definition}[$\fraction$-known state]\label{def:rou-known-state}
	A state $s$ is $\fraction$-known if there exist $\fraction A$ distinct actions such that the corresponding state-action pair $(s,a)$ is known.
\end{definition}

Intuitively, the idea of our proposed \walkname is: if the current state $s$ is not $\fraction$-known, choose an action $a$ that the agent has taken the most but $(s,a)$ is still unknown; if the current state is $\fraction$-known, select the action which most likely leads to a non-$\fraction$-known state. 
The agent also chooses actions randomly with a small probability $\beta$ to avoid being trapped in local optima.

Algorithm~\ref{alg:greedy_walk} shows the procedure of \textit{\walkname}, where $\mathsf{Random()}$ generates a random number from a uniform distribution in $[0,1]$; $n(s,a)$ is the total number of visits to state-action pair $(s,a)$; $n(s^\prime|s,a)$ is the total number of transitions from $s$ to $s^\prime$ by taking action $a$; the indicator function $\mathbb{I}(s^\prime \text{ is non-} \fraction \text{-known})$ is 1 if $s^\prime$ is not $\fraction$-known, and 0 otherwise.


\begin{algorithm}[!htbp]
\DontPrintSemicolon
\caption{$\beta$-CuriousWalking}
\label{alg:greedy_walk}
  \Input{The current state $s$, a hyper-parameter $\beta$}
  \Output{The chosen action $a^*$}
  \If{$\mathsf{Random}() < \beta$}{
    \KwRet{$a^* \gets $ a random action}
  }
  \eIf{$s$ is non-$\fraction$-known}{
    $\tilde{\actions} \gets \actions$ \;
    \ForEach{$a \in \tilde{\actions}$}{
      $\tilde{\actions} \gets \tilde{\actions} / \{a\}$ if (s,a) is known \;
    }
    $a^* \gets \argmax_a n(s,a), a \in \tilde{\actions}$
  }{
    \ForEach{$a \in \actions$}{
      $t(a) = \sum_{s^\prime} \frac{n(s^\prime|s,a)}{n(s,a)} \mathbb{I}(s^\prime \text{ is non-} \fraction \text{-known})$\;
    }
    $a^* \gets \argmax_a t(a), a \in \actions$
  }
  \KwRet{$a^*$}
\end{algorithm}


Compared with the balanced walking method used in both $E^3$ and RMax, \walkname: 
\begin{itemize}
	\item Encourages the agent to choose the actions it has the most experience with, until the action is known for the current state. So the agent rapidly knows $\fraction SA$ state-action pairs.
	\item Spreads the knowledge evenly, i.e., the agent attempts to know every state with $\fraction A$ actions instead of attempting to know all $A$ actions for some states but nothing for other states.
\end{itemize}

\subsection{\ourmod Algorithm}

The proposed \ourmod algorithm is described in Algorithm~\ref{alg:gwsm}. 
When there are less than $\fraction SA$ $m$-known state-action pairs, the agent keeps exploring with \walkname (see lines 7-14). 
As long as $\fraction SA$ pairs are $m$-known, the algorithm performs matrix completion for all \dmnames (see lines 15-20). 
$\mathsf{MatComp}$ could be any off-the-shelf matrix completion algorithm 
that solves the problem defined in Equation~\eqref{eq:mc_problem}. 
Note that matrix completion algorithms implicitly estimate the rank $r$ of the input matrix, so there is no need to specify the rank as an input to Algorithm~\ref{alg:gwsm}. 

\noindent \textbf{Known threshold.} $m$ is the least number of visits to one state-action pair to make the estimation and the completion accurate. The choice of $m$ is specified  in Theorem~\ref{thm:main}. 

\noindent \textbf{Fraction of known state-action pairs.} $\fraction$ controls the fraction of known state-action pairs, based on which the matrix completion can get convincing results for unknowns.
The value of $\fraction$ is determined by the structure of the underlying MDP. 
The more the states and actions in the MDP are interrelated, the smaller $\fraction$ can be. 
If the underlying MDP has completely unrelated dynamics, then $\fraction$ is set to be 1, and matrix completion does nothing but returning the empirical transition model itself. 

We will further discuss parameters $m$ and  $\fraction$ in theory and practice in the next two sections.
More importantly, we will show that the advantage of our proposed algorithm over previous model-based algorithms is larger for smaller $\fraction$. 
Even under the worst scenario of $\fraction=1$,  our \walkname improves learning efficiency under certain conditions.


\begin{algorithm}[!ht]
\DontPrintSemicolon
\caption{Greedy Inference Algorithm}
\label{alg:gwsm}
  \Input{$T, H, \epsilon, m, \fraction, \beta$}
  \Output{Near-optimal policy $\tilde{\pi}$ such that $V^{\tilde{\pi}} \geq V^* - \epsilon$}
  Initialize dynamic matrices $\{\admat^s\}_{s \in \states}, \admat^r$  \;
  Initialize $n(s,a),n(s^\prime|s,a),R(s,a)$ for all $s,s^\prime \in \states, a \in \actions$ \;
  Initialize $\mask$ as all zeros \;
  \For{episode $t\gets 1$ \KwTo $T$}{
  	$s_1 \gets$ initial state \;
  	\For{step $h\gets 1$ \KwTo $H$}{
      \eIf{$sum(\mask) < \fraction SA$}{
        $a_h \gets $ $\beta$-CuriousWalking($s_h, \beta$) \;
        Execute $a_h$, get $s_{h+1}$ and $r_{h+1}$ \;
        $n(s_h,a_h) \gets n(s_h,a_h) + 1$ \;
        $n(s_{h+1}|s_h,a_h) \gets n(s_{h+1}|s_h,a_h) + 1$ \;
        $R(s_h,a_h) \gets R(s_h,a_h) + r_{h+1}$ \;
        Update $\admat^{s_{h+1}}$ and $\admat^r$ by Equation~\eqref{eq:admat} \;
        \lIf{$n(s_h,a_h)\geq m$}{Update $\mask$}
        \If{$sum(\mask) \geq \fraction S A$}{
          \For{$s \in \states$}{
            $\rdmat^s \gets$ $\mathsf{\MC}$({$\admat^s,\mask$}) \;
          }
          $\rdmat^r \gets$ $\mathsf{\MC}$({$\admat^r,\mask$}) \;
          Set all state-action pairs as known \;
          Compute the optimal policy $\tilde{\pi}$ \;
        }
      }{
        Choose $a_h$ with optimal policy $\tilde{\pi}$ \;
      }
  	}
  }

\end{algorithm}

\subsection{\ourmod As a Framework}
\label{sec:framework}
Although Algorithm~\ref{alg:gwsm} estimates the dynamics by directly averaging collected samples, which is similar to the classic RMax algorithm,
\ourmod can also be regarded as a framework and can be combined with other model-based methods.
The simple RMax-style structure in Algorithm~\ref{alg:gwsm} is an illustration of how \ourmod could be combined with a model-based method; the analysis we will provide in Section~\ref{sec:analysis} exhibits how \ourmod could improve a model-based method. 

The key ideas of \ourmod are to infer the unknown dynamics as well as to know the environment greedily and evenly, which improve the model-based method combined.  
For example, in algorithms driven by confidence interval estimation, such as MBIE~\cite{strehl2005a}, UCRL2~\cite{jaksch2010near} and etc, we can also use matrix completion to recover the ``uncertain dynamics'' using the ``dynamics with high confidence''. 
This extra operation will not affect what has been learned, but rather make a guaranteed estimation of the unlearned parts. 
Therefore, the learning process is boosted by utilizing the internal structures of the environment, and as a result, samples are saved.

Overall, it is not our goal to propose a specific algorithm with the best complexity. Instead, we attempt to improve any model-based algorithm by inferring the dynamics.

\section{Theoretical Analysis on Complexities}
\label{sec:analysis}









In this section, we analyze the computational complexity, sample complexity and space complexity of \ourmod. By comparing with existing model-based methods, we show that \ourmod achieves a much better computational complexity and improves the sample complexity under mild conditions.

\subsection{Computational Complexity}
\label{ana:cc}

Theorem~\ref{thm:cc} states the computational complexity of \ourmod.

\begin{theorem}[Computational Complexity of \ourmod] 
\label{thm:cc}
Given an MDP $M$ with $S$ states and $A$ actions, if \ourmod is executed for $\mathcal{N}$ steps, then the total computational complexity of \ourmod is
\begin{equation}
	\mathcal{\widetilde{O}}(\varphi + S\max\{S,A\} + \mathcal{N}),
\end{equation}
where $\varphi$ be the number of computations for one dynamic programming operation, i.e., updating the policy by solving Bellman equations.
\end{theorem}
\noindent  \textbf{Remark.} $\varphi$ depends on the environment size. More specifically, if the maximum number of iterations for dynamic programming is set as $\mathcal{U}$, then $\varphi=\mathcal{O}(SA\mathcal{U})$. 

\paragraph{Proof.} During the execution of \ourmod, both dynamic programming and matrix completion are implemented only once, which lead to $\mathcal{O}(\varphi)$ and $\mathcal{\widetilde{O}}(S\max\{S,A\})$~\cite{gamarnik2017matrix} computations. For every time step, \ourmod updates $n(s,a), n(s^\prime|s,a)$ and $R(s,a)$, which can be done in constant time. Although \walkname in Algorithm~\ref{alg:greedy_walk} performs a loop over all the actions, in practice we are able to find the best action within constant time or logarithmic time through maintaining the known-ness table. These constant-time per step computations together lead to the $\mathcal{\widetilde{O}}(\mathcal{N})$ term.

\paragraph{Comparison with RMax and RTDP-RMAX\cite{strehl2012incremental}.} 
We first compare the cost of dynamic programming, the major computational burden for most RL algorithms.
Since within each dynamic programming, the amount of computation required is the same for \ourmod and other model-based methods, we use the number of dynamic programming operations as the metric for computation complexity comparison.

\begin{lemma}
The number of dynamic programming operations required by \ourmod is $\mathcal{O}(1)$, whereas the number of dynamic programming operations required by RMax and RTDP-RMAX are $\mathcal{O}(S)$ and $\mathcal{O}(\frac{SA\varepsilon}{V_{\max}})$.
\end{lemma}

RMax's computational complexity is $\mathcal{O}(S\varphi)$, as RMax computes the action values every time a new state is known. 
RTDP-RMAX is proposed to reduce the computational complexity of RMax. 
It initializes all action values to be $V_{\max}$, the maximum possible value for an episode, and only updates the value of one state-action pair when the value decreases more than some threshold $\varepsilon$. 
So it requires at most $\mathcal{O}(\frac{SA\varepsilon}{V_{\max}} \varphi)$ computations for the dynamic programming, where $\varepsilon$ is the error tolerance of action-value estimation.
Thus in terms of dynamic programming computation complexity, \ourmod is much faster than RMax and RTDP-RMAX. 

Besides the dynamic programming, as shown in Theorem~\ref{thm:cc}, GIM requires some constant-time operations per step, which is inevitable for all algorithms, and an extra matrix completion computation, which is a one-time cost and is negligible in a long learning process. 
Experiments in Section~\ref{sec:runtime} verifies this fact.

\subsection{Sample Complexity}
\label{ana:sc}
As stated in Section~\ref{sec:framework}, what we propose is a new exploration and estimation approach, that could be combined with model-based PAC algorithms \cite{strehl2008analysis,szita2010model,dann2015sample} to get lower sample complexity. 
In this section, we analyze the sample complexity of Algorithm~\ref{alg:gwsm} by adapting the analysis of RMax~\cite{kakade2003sample}. RMax is chosen due to its simplicity and versatility.

We now introduce a few notations that are essential in our analysis. 
(1) Denote the upper bounds of the condition number and the rank of every \dmname by $\kappa$ and $r$. 
(2) All \dmnames are at least $(\mu_0, \mu_1)$-incoherent.
(3) Let $\minsa=\min\{S,A\}$, and $\maxsa=\max\{S,A\}$. 

We also make the following two mild assumptions. 
\begin{assumption}\label{assump:small_diameter}
There is a known diameter $D$, such that any state $s^\prime$ is reachable from any state $s$ in at most $D$ steps on average. Assume that the diameter $D$ is smaller than the horizon $H$.
\end{assumption}
The assumption about diameter is commonly used in RL~\cite{jaksch2010near}, and it ensures the reachability of all states from any state on average. It is mild to assume $D < H$ as the horizon is often set large.

\begin{assumption}\label{assump:Gaussian_noise}
 The distribution of the estimation noise, $p(\cdot|\cdot,\cdot)-\hat{p}(\cdot|\cdot,\cdot)$ and $r(\cdot,\cdot)-\hat{r}(\cdot,\cdot)$, is sub-Gaussian with 0 mean. 
 And the estimation noises for different state-action pairs are independent.
\end{assumption}
This modeling of difference between the ground-truth probability and empirical estimation using sub-Gaussian variables is widely used.

The sample complexity of \ourmod is in Theorem~\ref{thm:main}.

\begin{theorem}[Sample Complexity of \ourmod] 
\label{thm:main}
	Given an MDP $M$ with fixed horizon $H$ and diameter $D$, suppose the upper bounds of the condition number, rank and incoherence parameters of \dmnames are $\kappa, r, \mu_0$ and $\mu_1$, for any $0<\epsilon<1$, $0 \leq \delta < 1$, with completion fraction 
	\begin{equation}
	\label{eq:rho}
		\begin{aligned}
		\fraction \geq 
		\Omega(\frac{1}{\sqrt{SA}}\kappa^2 \max\{ & \mu_0 r \sqrt{\frac{\maxsa}{\minsa}} \log \minsa, \\
		&\mu_0^2 r^2 \frac{\maxsa}{\minsa} \kappa^4, \mu_1^2 r^2 \frac{\maxsa}{\minsa} \kappa^4 \} ),
		\end{aligned}
	\end{equation}
	and the known threshold 
	\begin{equation}
	\label{eq:m}
		\begin{aligned}
		m \geq 
		\mathcal{O}(\frac{\kappa^4 r S H^2 \maxsa}{\fraction A \epsilon^2} ),
		\end{aligned}
	\end{equation}
	algorithm \ref{alg:gwsm} produces a policy $\hat{\pi}$, which satisfies $V^{\hat{\pi}}_M \geq V^*_M - \epsilon$ for all but 
	$
	\mathcal{O}(\frac{\kappa^4 r S^2 \maxsa H D}{ (1-\beta) \epsilon^2 } \log \frac{1}{\delta}) 
	$
	episodes, with probability at least $1-\delta-1/\minsa^3$.  
\end{theorem}
\noindent  \textbf{Remark.} 
For low-rank \dmnames, $\kappa$ tends to be small. 
In our experiments, $\kappa$ is typically less than 2. 
If we regard $\frac{\kappa^4 D}{(1-\beta)}$ as a constant, the sample complexity of \ourmod becomes $\mathcal{O}(\frac{r S^2 \maxsa H}{\epsilon^2} \log \frac{1}{\delta} )$.

\paragraph{Proof Sketch.} We first set $m$ to be the least number of visits to a state-action pair to make it known (see Condition~\ref{cond:m} in Appendix~\ref{app:known_th}). Then, we prove with at most $\mathcal{O}(\frac{\fraction mSAD}{(1-\beta)H} \log \frac{1}{\delta})$ episodes, we know $\fraction SA$ pairs as Lemma~\ref{lem:sc_m}. Finally, we prove the value of $m$ should be $\mathcal{O}(\frac{\kappa^4 r S H^2 \maxsa}{\fraction A \epsilon^2} )$ by Lemma~\ref{lem:sim_tmdp}, Lemma~\ref{lem:mc_bound} and Lemma~\ref{lem:pertur_bound}. The full proof and the lemmas are in Appendix~\ref{app:proof}.

\paragraph{Comparison with RMax.} 
The sample complexity of RMax, in our settings, is $\mathcal{O}(\frac{ S^2 A H^2}{\epsilon^3} \log \frac{1}{\delta} \log \frac{SA}{\delta})$\cite{kakade2003sample}. 
We compare the sample complexity of \ourmod and RMax in the following scenarios.\\
\noindent \textit{(1) When $A \geq S$.} \ourmod has lower sample complexity than RMax if 
$r < \mathcal{O}(\frac{H}{\epsilon} \log \frac{SA}{\delta})$.

\noindent \textit{(2) When $S>A$.} \ourmod has lower sample complexity than RMax if 
$r < \mathcal{O}(\frac{AH}{S\epsilon} \log \frac{SA}{\delta})$.

\noindent \textit{(3) Worst Scenario ($\fraction=1$).} We deactivate the matrix completion steps by simply setting $\fraction=1$ and $m=\mathcal{O}(\frac{ SH^2}{\epsilon^2} \log \frac{SA}{\delta})$, when the underlying MDP does not have any inner-related structure. This makes \ourmod follow \walkname until all the state-actions are known. In this case, the sample complexity becomes $\mathcal{O}(\frac{ S^2 A HD}{(1-\beta)\epsilon^2} \log \frac{1}{\delta} \log \frac{SA}{\delta})$. Because $H \gg D$ and $1-\beta$ is close to 1, \ourmod generates less non-$\epsilon$-optimal episodes than RMax does. So \walkname strategy itself saves samples. 

Note that $r\leq \min\{S,A\}$, so the conditions in (1) and (2) are satisfied for most tasks.

\paragraph{Achieve Lower Sample Complexity} 
\label{ana:lower_sc}
One may note that the sample complexity bound in Theorem~\ref{thm:main} is not optimal in terms of the dependency on $S$ and $A$. When $A>S$, \ourmod needs $\mathcal{\tilde{O}}(S^2 A)$ samples to learn a near-optimal policy. 
However, the best known bounds of model-based algorithms are of order $\tilde{\mathcal{O}}(SA)$~\cite{szita2010model,dann2017unifying,dann2018policy}. The saved factor of $S$ results from the direct analysis of the value function (and an imperfect model approximation). Since we claim that \ourmod can work as a framework, can \ourmod also achieve linear dependency?
In our analysis and the original analysis of RMax, the known threshold is at least $m=\mathcal{O}(S\ln S)$, but as indicated by~\cite{kearns1999finite},  with specific updating strategies, $m=\mathcal{O}(\ln S)$ samples might be enough to maximize the rewards. So it is possible for \ourmod to avoid an $S$ factor by incorporating re-estimating methods, although this is out of the scope of this paper.

\subsection{Space Complexity}
\label{ana:spc}
The memory \ourmod needs is mainly for the storage of \dmnames. Similar with other model-based RL algorithms, the space complexity of \ourmod is $\Theta(S^2 A)$, as we have $S+2$ matrices with size $S \times A$. 
In contrast, the space complexity of model-free algorithm such as Delayed Q-learning could be as low as $\mathcal{O}(SA)$. Although a large space complexity seems to be unavoidable for model-based methods, one can consider storing the sparse dynamics in a sparse format where only non-zero elements are stored if the dynamic matrices are sparse. Then the space complexity will be reduced to $\Theta(nnz)$, where $nnz$ is the number of non-zero entries.


\section{Experiments}
\label{sec:exp}

\subsection{Performance on Multiple Tasks}
\subsubsection{Tasks} 

To exhibit the universal applicability of \ourmod, we conduct experiments on multiple tasks of varying levels of hardness:
(1) \emph{Synthetic.} We create various MDPs by randomly generating the \dmnames with varying numbers of states, actions and ranks. 
(2) \emph{GridWorld.} A classic grid world task, with world size $4 \times 4$, slip probability 0.4 and step cost 0.2. 
(3) \emph{CasinoLand.} A challenging task constructed by \cite{strehl2004exploration}, which consists of six rooms and three levers. Pulling some levers may lead to a large reward with a small probability. 
(4) \emph{RiverSwim.} Another challenging task constructed by \cite{strehl2004exploration}, where a chain of 6 states represents a river, and an agent needs to ``swim'' from the first state to the last one to get a large reward. 
See Appendix~\ref{app:casinoland} for detailed description of CasinoLand and RiverSwim.

\subsubsection{Baselines} 

To verify the effectiveness and efficiency of \ourmod, it is compared against popular model-based and model-free methods: RMax, Q-learning, Delayed Q-learning, and Double Q-learning \cite{NIPS2010_3964} methods. We select RMax among all model-based methods, because Algorithm~\ref{alg:gwsm} is designed on the basis of RMax. So the effectiveness of our proposed strategies can be justified by comparing with RMax.
Note that it is also possible to apply similar strategies to other existing model-based algorithms, as claimed in Section~\ref{sec:framework}.

Moreover, we implement an ``optimal'' agent which knows all the dynamics and deploys the optimal policy from the beginning, as well as a ``random'' agent which chooses action randomly. 
The ``optimal'' agent is the best any agent could achieve, while the ``random'' agent is the worst any agent could perform.
We use the simple\_rl framework provided by~\citet{abel2019simplerl} to conduct the experiments.



\subsubsection{Reward Comparison}
\begin{figure}[!htbp]
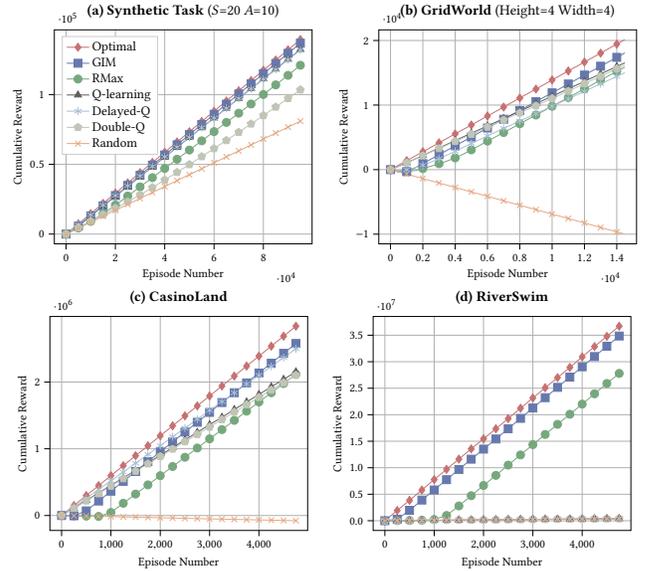

\centering
	\begin{subfigure}[t]{0.49\linewidth}
		\centering
		\input{\fighome/all/lowrank_S=20_A=10_r=2_runs=20.tex}  	
	\end{subfigure}
	\hfill
	\begin{subfigure}[t]{0.49\linewidth}
		\centering
		\input{\fighome/all/gridworld_h=4_w=4_runs=20.tex}	  
	\end{subfigure}
	\hfill
	\begin{subfigure}[t]{0.49\linewidth}
		\centering
		\input{\fighome/all/casinoland_runs=20.tex}	  
	\end{subfigure}
	\hfill
	\begin{subfigure}[t]{0.49\linewidth}
		\centering
		\input{\fighome/all/riverswim_runs=20.tex}	  
	\end{subfigure}
	\vspace{-1em}
  	\caption{Comparison of the mean cumulative reward over 20 runs of \ourmod and baselines on various tasks.} 
	\label{fig:all_acc}
	\vspace{-1em}
\end{figure}

We demonstrate the average cumulative reward over 20 runs of \ourmod and baselines on various tasks in Figure~\ref{fig:all_acc}. The plots are smoothed out by plotting every few hundred epochs for better illustration. Note that a line of reward increases linearly after it finds the best policy. Agents whose sample complexities are lower converge to the best policy earlier.
For hyper-parameters, we set the known threshold $m$ for both RMax and \ourmod to be 40, and set the completion fraction threshold $\fraction$ for \ourmod to be 0.8. We will discuss the influence of different settings on RMax and \ourmod later.
Among all these methods, \ourmod achieves the highest total reward. 
Although Q-learning, Delayed-Q and Double-Q converge to a good policy quickly, they are sometimes trapped in local-optima and cannot win in the long run. 
On the contrary, RMax figures out a policy that is near-optimal
, but it often takes more episodes to converge to that policy. 
\ourmod avoids these two drawbacks; it converges quickly and the returned policy is near-optimal.

\begin{table}[!htbp]
\vspace{-0.5em}
	\centering
	\tabcolsep=0.09cm
	\begin{tabular}{ c || c c| c c c}
				& \multicolumn{2}{c|}{\emph{Model-based}} & \multicolumn{3}{c}{\emph{Model-free}} \\ \cline{2-6}
		        & \ourmod  & RMax & Q & Delayed-Q & Double-Q  \\ \hline
			\emph{Synthetic} & 539.19 & 694.15 & 364.83 & 530.47 & 619.41  \\ \hline 
			\emph{Gridworld} & 11.75 & 13.94 & 12.68 & 11.2 & 13.7  \\ \hline 
			\emph{Casinoland} & 7.21 & 8.18 & 3.61 & 3.00 & 3.93  \\ \hline 
			\emph{RiverSwim} & 7.62 & 6.06 & 5.87 & 4.88 & 6.33   
	\end{tabular}
	\caption{Comparison of running times in seconds.}
	\label{tbl:runtimes}
	\vspace{-2.5em}
\end{table}
\subsubsection{Running Time Comparison}
\label{sec:runtime}
Table~\ref{tbl:runtimes} provides the average running times of each agent on various tasks. 
In general, our model-based \ourmod is faster than model-based RMax, and is even comparable to the model-free methods which are generally faster as there is no need to maintain a model. 
For RiverSwim, \ourmod is slightly slower than RMax as the one-time cost of computation of matrix completion slows down the \ourmod agent. 
This one-time cost is less significant for time-consuming tasks where running time is the bottleneck. 

\subsubsection{Scale Up to Larger Environments}

\begin{figure}[!htbp]
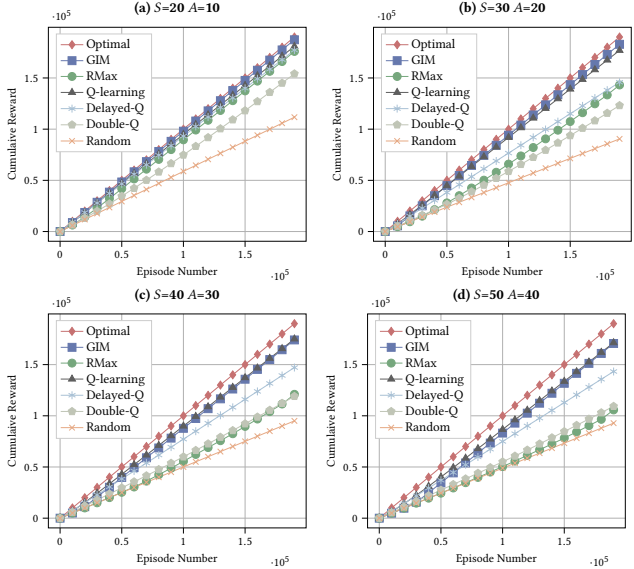

\vspace{-1em}
\centering
	\begin{subfigure}[t]{0.49\linewidth}
		\centering
		\input{\fighome/more/lowrank_S=20_A=10_r=2_runs=20.tex}  	
	\end{subfigure}
	\hfill
	\begin{subfigure}[t]{0.49\linewidth}
		\centering
		\input{\fighome/more/lowrank_S=30_A=20_r=2_runs=20.tex}	  
	\end{subfigure}
	\hfill
	\begin{subfigure}[t]{0.49\linewidth}
		\centering
		\input{\fighome/more/lowrank_S=40_A=30_r=2_runs=20.tex}	  
	\end{subfigure}
	\hfill
	\begin{subfigure}[t]{0.49\linewidth}
		\centering
		\input{\fighome/more/lowrank_S=50_A=40_r=2_runs=20.tex}	  
	\end{subfigure}
	\vspace{-1em}
  	\caption{Comparison of the mean cumulative reward of \ourmod and baselines on synthetic tasks with various environment size.} 
	\label{fig:more_acc}
\end{figure}

Although many model-based algorithms perform well in small environments, it is usually not easy for them to work on large-scale problems. 
However, \ourmod avoids exhaustedly visiting and estimating the whole environment by inferring some dynamics. 
When the environment is highly internal-dependent, only the knowledge of a small fraction of the space is needed. 
Thus fewer samples and computations are consumed. 
To evaluate the scalability of \ourmod compared with baselines, we gradually enlarge the size of the synthetic task and show the cumulative rewards in Figure~\ref{fig:more_acc}. 
\ourmod performs well compared with baselines, while RMax fails to converge to the optima within the given number of episodes. 
We set the known thresholds for both \ourmod and RMax to be 100. 
As a result, 
the performance of learned policy by \ourmod is slightly worse than Q-learning, since $m=100$ is not adequate for estimating the large environments.
But one can anticipate higher rewards by \ourmod, once $m$ is set higher.

The running times corresponding to Figure~\ref{fig:more_acc}, are shown in Table~\ref{tbl:runtimes_more}, where one can find \ourmod is much faster than RMax and Double Q-learning. 
As the environment size gets larger, \ourmod takes slightly longer to run, while RMax spends much more time due to the increased computation requirements.
\vspace{-1em}
\begin{table}[!htbp]
\vspace{-1em}
	\centering
	\tabcolsep=0.12cm
	\vspace{-2em}
	\begin{tabular}{ c || c c| c c c}
				& \multicolumn{2}{c|}{\emph{Model-based}} & \multicolumn{3}{c}{\emph{Model-free}} \\ \cline{2-6}
		        & \ourmod  & RMax & Q & Delayed-Q & Double-Q  \\ \hline
			\emph{S=20, A=10}  & 2165  & 2410  & 1386  & 2244  & 2465 \\ \hline
			\emph{S=30, A=20}  & 2296  & 4581  & 1635  & 2541  & 3107 \\ \hline
			\emph{S=40, A=30}  & 2304  & 9778  & 2007  & 2477  & 3413 \\ \hline
			\emph{S=50, A=40}  & 2423  & 19870  & 2287  & 2399  & 3424 
	\end{tabular}
	\caption{Comparison of running times (in seconds) on synthetic tasks with various environment size. The number of episodes are all set to be $2\times 10^5$.}
	\label{tbl:runtimes_more}
	\vspace{-2em}
\end{table}
\vspace{-1em}

\subsection{Experiments on Parameters}
In practice, it is important to select an appropriate known threshold $m$ for \ourmod and RMax. \ourmod also requires a completion fraction $\fraction$. Although we can choose $\fraction$ based on experience, it is related to the properties of the \dmnames ($\kappa, r, \mu_0, \mu_1$) as proved in Theorem~\ref{thm:main}.
So, we design and conduct a series of systematic tests to study how $m, \kappa, r, \mu_0, \mu_1$ (for fixed $\fraction$) influence the learning effectiveness and efficiency of \ourmod, in comparison with RMax.

The following three measurements are evaluated. (1) \textbf{\avgrew}: average reward per episode; (2) \textbf{\numepi}: number of episodes needed to know all states and actions; and (3) \textbf{\pavgrew}: average reward after knowing all states and actions (after exploration), which reflects accurateness of the learned dynamics. 

\subsubsection{Values of the Known Threshold $m$}

\begin{figure}[!ht]
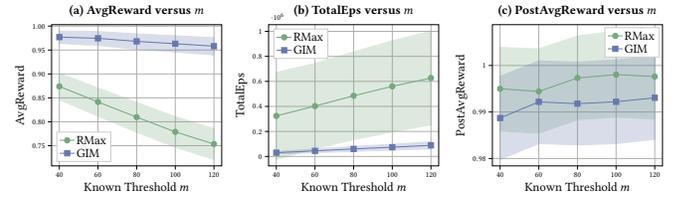

\centering
\vspace{-1em}
	\begin{subfigure}[t]{0.3\columnwidth}
		\centering
		\input{\fighome/avg/lowrank_S=20_A=10_r=2_runs=20.tex}	  	
	  	\label{sfig:com_avg_lowrank}
	\end{subfigure}
	\hfill
	\begin{subfigure}[t]{0.3\columnwidth}
		\centering
		\input{\fighome/times/lowrank_S=20_A=10_r=2_runs=20.tex}	  
	  	\label{sfig:com_times_lowrank}
	\end{subfigure}
	\hfill
	\begin{subfigure}[t]{0.3\columnwidth}
		\centering
		\input{\fighome/pavg/lowrank_S=20_A=10_r=2_runs=20.tex}
	  	\label{sfig:com_pavg_lowrank}
	\end{subfigure}
	\vspace{-2em}
  	\caption{Comparison of \avgrew, \numepi, and \pavgrew of RMax and \ourmod on synthetic task with different known thresholds. S=20, A=10 and rank=2.} 
	\label{fig:th_compare}
	\vspace{-1.5em}
\end{figure}

We run experiments on synthetic and GridWorld tasks with different known threshold $m$. Figure~\ref{fig:th_compare} shows the \avgrew, \pavgrew and \numepi of RMax and \ourmod with different values of $m$. 
For the same $m$, \ourmod gains more rewards (Figure~\ref{fig:th_compare}a), and completes exploration faster than RMax (Figure~\ref{fig:th_compare}b), with a slightly worse returned policy (Figure~\ref{fig:th_compare}c). When $m$ varies, RMax requires much more episodes to explore, while \ourmod is more robust to the changing $m$ as shown in Figure~\ref{fig:th_compare}b.
We present the results on higher-rank synthetic tasks and the GridWorld task in Appendix~\ref{app:exp_th}.

\subsubsection{Properties of Dynamic Matrices}

\begin{figure}[!htbp]
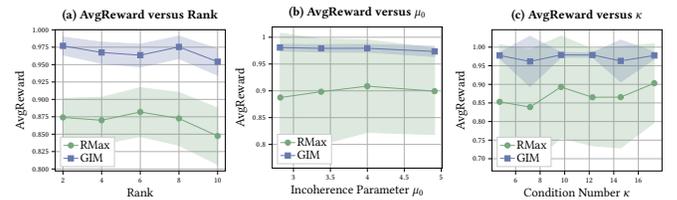

\centering
\vspace{-1em}
	\begin{subfigure}[t]{0.3\linewidth}
		\centering
		\input{\fighome/rank/ranks_avg_shade_runs=20.tex}	  
	  	\label{sfig:rank_avg}
	\end{subfigure}
	\hfill
	\begin{subfigure}[t]{0.3\linewidth}
		\centering
		\input{\fighome/mus/lines/admu_rew_line_th40_mu0.tex}	  
	\end{subfigure}
	\hfill
	\begin{subfigure}[t]{0.3\linewidth}
		\centering
		\input{\fighome/mus/lines/admu_rew_line_th40_kappa_runs=20.tex}	  
	\end{subfigure}
	 \vspace{-2em}
  	\caption{Comparison of \avgrew of \ourmod and RMax 
	with varying ranks, $\mu_0$ and $\kappa$. $S=20$, $A=10$ and $m=40$. The generated ranks for \textbf{(b)} and \textbf{(c)} are 2 and 4 respectively. } 
	\label{fig:spectral}
	\vspace{-1.5em}
\end{figure}

We visualize the \avgrew of \ourmod and RMax under varying dynamic matrix ranks $r$'s, incoherence parameter $\mu_0$'s and condition number $\kappa$'s in Figure~\ref{fig:spectral}.

\noindent \textit{(1) Influence of $r$.}  As Figure~\ref{fig:spectral}a shows, the average rewards of both \ourmod and RMax drop \textbf{slightly} when rank becomes higher. Even if the dynamic matrix is full-rank, the policy returned by \ourmod still obtains a high reward.

\noindent \textit{(2) Influence of $\mu_0$ and $\kappa$.} We see from Figure~\ref{fig:spectral}b and Figure~\ref{fig:spectral}c that \ourmod outperforms RMax, and does not change much with varying $\mu_0$ or $\kappa$.

Experiments of matrix properties are conducted on synthetic tasks because it is easy to control the properties of their underlying MDPs. But the observed connections between studied properties and learning results can be extended to any other tasks. Additional experimental results including \pavgrew, \numepi are shown in Appendix~\ref{app:property}, as well as the results for $\mu_1$. 

\section{Conclusion}
\label{sec:conc}
This paper proposes a model-based RL algorithm called \ourmod. \ourmod utilizes the internal structures of MDPs to infer the unknown dynamics, and uses a novel exploration strategy to efficiently explore the environment.
Theoretical analysis and empirical results show that \ourmod can reduce both sample complexity and computational complexity when combined with classic model-based algorithms.
We envision incorporating our proposed techniques for multi-task RL where internal structures of MDPs can connect different tasks.


\bibliographystyle{ACM-Reference-Format}  

\newpage
\onecolumn
\appendix

{\begin{center}\textbf{\LARGE Appendix: \mytitle}\end{center}}

\section{Matrix Completion and Incoherence Property}
\label{app:mc}

\noindent \textbf{Matrix Completion}

The matrix completion problem, which considers the recovery of some missing entries from random sampling of the matrix, is widely-used in many practical fields like recommendation, information diffusion and etc. 

A typical matrix completion problem can be stated as follows. We have a numerical matrix $\mymatrix{M}$ with $n_1$ rows and $n_2$ columns. If we only observe $m$ entries where $m < n_1 n_2$, are we able to recover the remaining unknown entries accurately and efficiently?
Apparently, we cannot find the true recovery without any restriction and additional information. The matrices which we can complete from partial entries must have some certain kinds of structure. One important property is the low-rankness, i.e., the matrix should be low-rank or approximately low-rank. An $n_1 \times n_2$ matrix $\mymatrix{M}$ with rank $r$ can be represented by $n_1 n_2$ numbers, but it only has $(n_1+n_2-r)r$ degrees of freedom (can be derived through counting parameters of SVD). So it is unnecessary to known all the $n_1 n_2$ entries to determine the matrix $\mymatrix{M}$. When $r$ is small, theoretically, one can uniquely determine the whole matrix with only a small fraction of the matrix.

\noindent \textbf{Incoherence Conditions}

However, not any arbitrary low-rank matrix can be recovered from a few entries. For example, the following rank-1 matrix only has one non-zero entry ($M_{11}=1$), so it is impossible to know the matrix exactly unless all of the entries have been observed. 
$$
\mymatrix{M} = 
\begin{bmatrix}
1 & 0 & \cdots & 0 & 0 \\
0 & 0 & \cdots & 0 & 0 \\
\vdots & \vdots & \vdots & \vdots & \vdots \\
0 & 0 & \cdots & 0 & 0
\end{bmatrix}
$$ 

Intuitively, the information of the matrix cannot be too concentrated in some rows or columns. And its singular vectors should be spread instead of concentrated, that is, uncorrelated with the standard basis. Thus we have the so-called \textit{standard incoherence}. 

Suppose the rank-$r$ SVD of $\mymatrix{M}$ is $\mymatrix{U}\mymatrix{\Sigma}\mymatrix{V}\tran$, then $\mymatrix{M}$ is said to satisfy the \textit{standard incoherence condition} with parameter $\mu_0$ if
\begin{align}
\max_{1\leq i \leq n_1} \| \mymatrix{U}\tran \myvector{e}_i \|_2 \leq \mu_0 \sqrt{\frac{r}{n_1}} \\
\max_{1\leq j \leq n_2} \| \mymatrix{V}\tran \myvector{e}_j \|_2 \leq \mu_0 \sqrt{\frac{r}{n_2}}
\end{align}
where $\myvector{e}_i$ are the $i$-th standard basis with appropriate dimension. The smallest possible $\mu_0$ for any matrix can be 1, when the singular vectors all have equal magnitudes. And it is easy to get $\mu_0 \leq \max\{n_1,n_2\} / r$. The more concentrated the singular vectors are, the higher $\mu_0$ is.

To get guarantees for all values of the rank, \cite{candes2009power} proposes another incoherence condition called \textit{strong incoherence condition}. A matrix $\mymatrix{M}$ satisfies the \textit{strong incoherence condition} with parameter $\mu_1$ if 
\begin{align}
\max_{i,j} | (\mymatrix{U}\mymatrix{V}\tran)_{ij} |  \leq \mu_1 \sqrt{\frac{r}{n_1 n_2}}.
\end{align}
The \textit{strong incoherence condition} requires the left and right singular vectors of the matrix to be unaligned with each other. While a paper \cite{chen2013incoherence} has shown that this condition is not necessary for matrix completion without noise.

In our model, we perform matrix completion on the dynamic matrix, which is the transition probabilities to a specific state from all state-actions pairs. In order to get higher recovery accuracy with less known state-action pairs,  it is required that (1) the matrix is low-rank, which is true when different actions/states have the same transition probabilityies to the same state. (2) the transitions to a specific state are not highly concentrated on some state-action pairs.

\section{Proof Details of Theorem~\ref{thm:main}}
\label{app:proof}










\subsection{Sample Complexity with Known Threshold \textbf{m}}
\label{app:known_th}

As in RMax, let $m$ be the number of visits needed to make a state-action pair known, and we assume the following condition holds for $m$:
\begin{condition}
\label{cond:m}
	For a given MDP $M$, the known threshold $m$ is chosen such that the optimal policy $\hat{\pi}$ for the completed MDP $\widetilde{M}$ satisfys, with probability at least $1-\delta$
	\begin{equation}
	\label{eq:m_cond_all}
		| V^{\tilde{\pi}}_M - V^{\tilde{\pi}}_{\widetilde{M}} | \leq \epsilon
	\end{equation}
\end{condition}

Then with the parameter $m$, we can express the sample complexity as

\begin{lemma}[Sample Complexity in terms of $m$]
\label{lem:sc_m}
	Let $M$ be an MDP with fixed horizon $H$. If $\tilde{\pi}$ is the policy computed by \ourmod, then for any starting state $s_0$, with probability at least $1-2\delta$, we have $V_M^{\tilde{\pi}} \geq V_M^* - 2 \epsilon$ for all but $\mathcal{O}(\frac{\fraction mSAD}{(1-\beta)H} \log \frac{1}{\delta})$
\end{lemma}
\textbf{Remark.} $\beta$ is the probability of randomly selecting action, which can be set by hand. 

\textbf{Compare with RMax.} 
For Rmax algorithm, the sample complexity is $\mathcal{O}(\frac{mSA}{\epsilon} \log \frac{1}{\delta})$.





\begin{proof} (of Lemma~\ref{lem:sc_m}) 
Because of Condition~\ref{cond:m}, i.e., $| V^{\tilde{\pi}}_M - V^{\tilde{\pi}}_{\widetilde{M}} | \leq \epsilon$ with probability at least $1-\delta$, we have
\begin{align*}
	V^{\tilde{\pi}}_M & \geq V^{\tilde{\pi}}_{\widetilde{M}} - \epsilon \\
	& \geq V^{\pi^*_M}_{\widetilde{M}} - \epsilon \\
	& \geq V^{\tilde{\pi}}_{\widetilde{M}} - 2 \epsilon.
\end{align*}

So, after completion, the required value can immediately be reached. We only need to bound the number of exploration episodes.

Due to the \walkname strategy, in every step, one of the following three cases will happen: (1) with probability $\beta$, a random action is selected. (2) with probability $1-\beta$, visit an unknown state-action pair. (3) with probability $1-\beta$, choose an action which has the largest probability to a non-$\fraction$-known state.  

Note that the exploration ends when there are $\fraction SA$ known state-action pairs, and $m$ visits are needed to make an unknown pair known. For case (2), a visit to unknown pair happens. For case (1) and (3), the agent starts a trajectory going to a non-$\fraction$-known state, and then either (1) or (2) happens. Thus, with probability $1-\beta$, there is a successful visit to an unknown pair within at most $D$ steps. 

Further, in $N$ steps, the expected number of visits to unknown pairs is $(1-\beta)N/D$. Then, using Hoeffding's inequality, we can get

$$
N = \mathcal{O}\big( \frac{\fraction mSAD}{1-\beta} \log \frac{1}{\delta} \big)
$$

steps are enough to have $\fraction m SA$ visits to unknown pairs with probability at least $1-\delta$.

Therefore, the number of exploration episodes is bounded by $\mathcal{O}\big( \frac{\fraction mSAD}{(1-\beta)H} \log \frac{1}{\delta} \big)$, with probability at least $1-\delta$. Note that the order of the result will not be changed even if the agent is reset in the beginning of every episode and has to restart a trajectory, since we assume $H \gg D$.

\end{proof}



In order to prove Theorem~\ref{thm:main}, we need the following two lemmas: simulation lemma and perturbation bound lemma.

\begin{lemma}[Simulation Lemma]
\label{lem:sim_tmdp}
	Given two MDPs $M$ and $M^\prime$ in the same state-action space with the same horizon $H$, and their corresponding dynamics $p_M(\cdot|\cdot,\cdot), r_M(\cdot,\cdot)$ and $p_{M^\prime}(\cdot|\cdot,\cdot), r_{M^\prime}(\cdot,\cdot)$, if for any state $s$ and action $a$,
	\begin{align*}
		\| p_M(\cdot|s,a) - p_{M^\prime}(\cdot|s,a)\|_1 &\leq \epsilon, \text{ and } \\
		 | r_M(s,a) - r_{M^\prime}(s,a) | &\leq \epsilon
	\end{align*} 
	then for any policy $\pi$, we have $| V^\pi_M - V^\pi_{M^\prime} | \leq (H+1) \epsilon$.
\end{lemma}

\begin{proof} (of Lemma~\ref{lem:sim_tmdp}) 


We let $\tau$ denote a trajectory in MDP, and $\mathcal {T}_h$ the set of all trajectories of length $h$, $P^\pi_M(\tau)$ the probability of observing trajectory $\tau$ in MDP $M$ following policy $\pi$, and $U_M(\tau)$ the expected average reward of going through trajectory $\tau$ in MDP $M$. 

Then the difference between the values of two MDPs under policy $\pi$ is
(For simplixity, we omit the superscript $\pi$ for all variables.)

\begin{align*}
 | V_M - V_{M^\prime} | 
 &= \left| \sum _ { \tau \in \mathcal { T } _ { H } } \left[ P _ { M } ( \tau ) U _ { M } ( \tau ) - P _ { M ^ { \prime } } ( \tau ) U _ { M ^ { \prime } } ( \tau ) \right] \right| \\
 &\leq \left| \sum _ { \tau \in \mathcal { T } _ { H } } \left[ P _ { M } ( \tau ) U _ { M } ( \tau ) - P _ { M } ( \tau ) U _ { M ^ { \prime } } ( \tau ) + P _ { M }  ( \tau ) U _ { M ^ { \prime } } ( \tau ) - P _ { M ^ { \prime } }   ( \tau ) U _ { M ^ { \prime } } ( \tau ) \right] \right| \\
 &\leq \left| \sum _ { \tau \in \mathcal { T } _ { H } } \left[ P _ { M } ( \tau ) \left( U _ { M } ( \tau ) - U _ { M ^ { \prime } } ( \tau ) \right) \right] \right| + \left| \sum _ { \tau \in \mathcal { T } _ { H } } \left[ U _ { M ^ { \prime } } ( \tau ) \left( P _ { M } ( \tau ) - P _ { M ^ { \prime } }  ( \tau ) \right) \right] \right| \\
 &\leq  \epsilon \left| \sum _ { \tau \in \mathcal { T } _ { H } } P _ { M }  ( \tau ) \right| + \left| \sum _ { \tau \in \mathcal { T } _ { H } } \left[ P _ { M } ( \tau ) - P _ { M ^ { \prime } } ( \tau ) \right] \right| \\
 &=  \epsilon + \left| \sum _ { \tau \in \mathcal { T } _ { H } } \left[ P _ { M } ( \tau ) - P _ { M ^ { \prime } } ( \tau ) \right] \right| \\
\end{align*}

The bound of the second term is given by

\begin{align*}
& \left| \sum _ { \tau \in \mathcal { T } _ { H } } \left[ P _ { M } ( \tau ) - P _ { M ^ { \prime } } ( \tau ) \right] \right| \\
=& \left| \sum_{\tau^\prime \in \mathcal{T}_{H-1}} \sum_{s^\prime} P_M(\tau^\prime) P_M(s^\prime|\tau^\prime) - P_{M^\prime}(\tau^\prime) P_{M^\prime}(s^\prime|\tau^\prime) \right| \\
\leq& \left| \sum_{\tau^\prime \in \mathcal{T}_{H-1}, s^\prime} P_M(\tau^\prime) P_M(s^\prime|\tau^\prime) - P_{M^\prime}(\tau^\prime) P_M(s^\prime|\tau^\prime) \right| + \left| \sum_{\tau^\prime \in \mathcal{T}_{H-1}, s^\prime} P_{M^\prime}(\tau^\prime) P_M(s^\prime|\tau^\prime) - P_{M^\prime}(\tau^\prime) P_{M^\prime}(s^\prime|\tau^\prime) \right| \\
=& \left| \sum_{\tau^\prime \in \mathcal{T}_{H-1}} P_M(\tau^\prime) - P_{M^\prime}(\tau^\prime) \right| \sum_{s^\prime} P_M(s^\prime|\tau^\prime) + \sum_{\tau^\prime \in \mathcal{T}_{H-1}} P_{M^\prime}(\tau^\prime) \left| \sum_{s^\prime} P_M(s^\prime|\tau^\prime) - P_{M^\prime}(s^\prime|\tau^\prime) \right| \\
\leq& \left| \sum_{\tau^\prime \in \mathcal{T}_{H-1}} P_M(\tau^\prime) - P_{M^\prime}(\tau^\prime) \right| +  \epsilon
\end{align*}

So by recursing, we can get the final bound over the second term, and then the bound of Lemma~\ref{lem:sim_tmdp} holds.

\end{proof}

\begin{lemma}[Perturbation Bound of Dynamic Matrices Completion]
\label{lem:pertur_bound}
	Let $\dmatset$ be the groud truth \dmnames for a specific RL task, $\admatset$ the empirical \dmnames for the same task, and $\mask$ the knownness mask. Let $\{\noisem^k\}_{k=1}^{S+1}$ be the noise matrices, 
	so $\mask \odot \dmat^k = \mask \odot \admat^k + \mask \odot \noisem^k$ for $k=\krange$.
	If the fraction of observed entries $\fraction$ satisfies
	\begin{equation}
		\begin{aligned}
		\fraction \geq \Omega(\frac{1}{\sqrt{SA}}\kappa^2 \max\{ \mu_0 r \sqrt{\frac{\maxsa}{\minsa}}, \mu_0^2 r^2 \frac{\maxsa}{\minsa} \kappa^4, \mu_1^2 r^2 \frac{\maxsa}{\minsa} \kappa^4 \} ),
		\end{aligned}
	\end{equation}
	and if 
	then with probability at least $1-1/\minsa^3$, for any $k=\krange$, the result \dmname $\rdmat^k$ returned by the function \textit{MatrixCompletion}$(\admat^k, r)$ in \ref{alg:gwsm} satisfys 
	\begin{equation}
	\| \rdmat^k - \dmat^k \|_{\text{max}} \leq \mathcal{O} \Big( \kappa^2 \sigma \sqrt{\frac{r \maxsa}{\fraction SA}}\Big),
	\end{equation}
	where $\kappa$ and $r$ are respectively the supremum of the condition number and the rank of all dynamic matrices; $\sigma$ is the maximum variance proxy for noisy entries.
\end{lemma}

To prove Lemma~\ref{lem:pertur_bound}, we first bound the perturbation of matrix completion. There are many existing works about matrix completion, and since the specific matrix completion algorithms are not our focus in this paper, we simply choose a theoretical bound stated by Theorem 1.2 of paper \cite{keshavan2010matrix} and reorganize it as Lemma~\ref{lem:mc_bound}. Note that a tighter bound can be achieved with other algorithms and theories. 
\begin{lemma}
\label{lem:mc_bound}
	Let $\mymatrix{M} \in \mathbb{R}^{m \times n}$ (suppose $m \geq n$) be a $(\mu_0, \mu_1)$-incoherent matrix with rank-$r$ and condition number $\kappa$. $K$ is a subset of the complete set of entries $[n_1] \times [n_2]$. If we observe $\mymatrix{P}^K \odot \hat{\mymatrix{M}} = \mymatrix{P}^K \odot \mymatrix{M} + \mymatrix{P}^K \odot \mymatrix{Z}$, where $\mymatrix{Z}$ is a noise matrix, whose entries are independent random variables, with zero mean and sub-gaussian tails with variance proxy $\sigma^2$.
	Then with probability $1-1/n^3$, the completed matrix $\tilde{\mymatrix{M}}$ satisfys 
	$$
	\frac{1}{\sqrt{mn}}\| \mymatrix{M} - \tilde{\mymatrix{M}} \|_F \leq \mathcal{O} \Big( \kappa^2 \sigma \sqrt{\frac{rm}{|K|}} \Big),
	$$
	provided that
	$$
	|K| \geq \Omega(\sqrt{mn}\kappa^2 \max\{ \mu_0 r \sqrt{\frac{m}{n}} \log n, \mu_0^2 r^2 \frac{m}{n} \kappa^4, \mu_1^2 r^2 \frac{m}{n} \kappa^4 \} )
	$$
\end{lemma}

The proof of this lemma is in the third section of paper \cite{keshavan2010matrix}. Now we can proceed to prove Lemma~\ref{lem:pertur_bound}.

\begin{proof} (of Lemma~\ref{lem:pertur_bound}) 

We first analyze the $k$-th dynamic matrix $\dmat^k$. Let $r_k$, $\kappa_k$ and $\sigma_k$ denote the rank, the condition number, and the maximum absolute entry value of $\dmat^k$ respectively. 


Because of the sub-Gaussian assumption, the entries of $\noisem^k$ are independent random variables with zero mean, and sub-Gaussian tails with variance proxy $\sigma_k^2 \leq \epsilon^2_0$.
Then we have 
\begin{align*}
	\frac{1}{\sqrt{SA}}\| \dmat^k - \rdmat^k \|_F \leq \mathcal{O} \Big( \kappa_k^2 \sigma_k \sqrt{\frac{r_k \maxsa}{\fraction SA}}\Big).
\end{align*}




Sum over all \dmnames, we get
\begin{align*}
	\frac{1}{\sqrt{SA}} \sum_{k=1}^{S+1}\| \dmat^k - \rdmat^k \|_F \leq \mathcal{O} \Big( \sum_{k=1}^{S+1} \kappa_k^2 \sigma_k \sqrt{\frac{r_k \maxsa}{\fraction SA}}\Big).
\end{align*}

The above inequality is equivalent to
\begin{align*}
	\frac{1}{\sqrt{SA(S+1)}} \sqrt{\sum_{k=1}^{S+1} \sum_{i=1}^{S} \sum_{j=1}^{A} (\dmats^k_{ij} - \rdmats^k_{ij})^2 } \leq \mathcal{O} \Big(\kappa^2 \sigma \sqrt{\frac{r \maxsa}{\fraction SA}}\Big),
\end{align*}
where $r = \max \{ r_k |k=\krange \}$, $\kappa = \max \{\kappa_k | k=\krange \}$ and $\sigma = \max \{ \sigma_k | k=\krange \}$.

The LHS expression is the mean RMSE. Then we can extract every single state-action pair's dynamics and evaluate their $\ell_2$ losses as

\begin{align*}
	\frac{1}{\sqrt{S+1}} \sqrt{\sum_{k=1}^{S+1} (\dmats^k_{ij} - \rdmats^k_{ij})^2 } \leq \mathcal{O} \Big( \kappa^2 \sigma \sqrt{\frac{r \maxsa}{\fraction SA}}\Big), \forall i,j.
\end{align*}

It implies that for any state $s$ and action $a$, we have:

\begin{align*}
	\sqrt{\sum_{k=1}^{S+1} (p(s_k|s,a) - \hat{p}(s_k|s,a))^2 + (r(s,a) - \hat{r}(s,a))^2 } \leq \mathcal{O} \Big( \sqrt{S+1} \kappa^2 \sigma \sqrt{\frac{r \maxsa}{\fraction SA}}\Big)
\end{align*}

Because of the equivalence of norms, the sum of absolute values of dynamic differences (a.k.a. the $\ell_1$ norm) satisfies

\begin{align*}
	\sum_{k=1}^{S+1} |p(s_k|s,a) - \hat{p}(s_k|s,a)| + |r(s,a) - \hat{r}(s,a)| \leq \mathcal{O} \Big( (S+1) \kappa^2 \sigma \sqrt{\frac{r \maxsa}{\fraction SA}}\Big).
\end{align*}

Then we get 
\begin{align*}
	\|p(\cdot|s,a) - \hat{p}(\cdot|s,a)\|_1 &\leq \mathcal{O} \Big( (S+1) \kappa^2 \sigma \sqrt{\frac{r \maxsa}{\fraction SA}}\Big), \\
	|r(s,a) - \hat{r}(s,a)| &\leq \mathcal{O} \Big( (S+1) \kappa^2 \sigma \sqrt{\frac{r \maxsa}{\fraction SA}}\Big).
\end{align*}

\end{proof}

\subsection{Proof of Theorem~\ref{thm:main}}

Now we can proceed to prove the main theorem of sample complexity.

\textbf{Setting the known threshold $\boldsymbol{m}$.}
\label{sec:set_m}
In Lemma~\ref{lem:sc_m}, we assume that we know a good known threshold $m$ such that the estimated values of the completed \dmnames can achieve a near-optimal value (Condition~\ref{cond:m} holds). 

For similicity, we define an operator $dist(\cdot,\cdot)$ to measure the distance of two MDPs as 
$$dist(M, M^\prime) = \max_{i,j}\{ \max \{\|p_M(\cdot|s_i,a_j) - p_{M^\prime}(\cdot|s_i,a_j)\|_1, |r_M(s_i,a_j) - r_{M^\prime}(s_i,a_j)|\} \}$$


Now we discuss the appropriate value for $m$ in order to satisfy Condition~\ref{cond:m}.



We know that Condition~\ref{cond:m} is true when the following equation holds.
\begin{equation}
	| V^\pi_{M} - V^\pi_{\widetilde{M}} | \leq \epsilon/2
\end{equation}

Based on Lemma~\ref{lem:sim_tmdp}, the above equation can be satisfied when
\begin{equation}
\label{eq:sim}
dist(M, \widetilde{M}) \leq \frac{\epsilon}{H+1}, \forall k=\krange
\end{equation}

Lemma~\ref{lem:pertur_bound} shows that, when the known fraction is greater than some threshold, we have
\begin{equation}
\label{eq:pertu}
dist(M, \widetilde{M}) \leq \mathcal{O} \Big( (S+1) \kappa^2 \sigma \sqrt{\frac{r \maxsa}{\fraction SA}} \Big), \forall k=\krange
\end{equation}
with probability $1-1/\minsa^3$, where $\kappa$ and $r$ are respectively the supremum of the condition number and the rank of all dynamic matrices; $\sigma$ is the maximum variance proxy for noisy entries.









For any state-action pair $(s,a)$ and any transit-in state $s^\prime$, let $\hat{p}(s^\prime | s,a)$ be the estimated transition probability $\frac{n(s^\prime|s,a)}{n(s,a)}$. For simplicity, let $p$ denote $p(s^\prime | s,a)$ and $\hat{p}$ denote $\hat{p}(s | s,a)$,
Using Hoeffding's bound, we get
\begin{equation}
\mathbb{P}(|\hat{p}-p| \geq t) \leq 2 \exp (- 2 m t^2)
\end{equation}

So the entries of $\noisem$ is $\sqrt{1 / 4m }$-subguassian.

Then let the RHS of Equation~\ref{eq:pertu} less than $\epsilon / 2(S+1)H$, we get the result that with probability $1-1/\minsa^3$
\begin{equation}
	m \geq \mathcal{O}(\frac{\kappa^4 r S H^2 \maxsa}{\fraction A \epsilon^2} )
\end{equation}

then Condition~\ref{cond:m} holds.

\textbf{Combining the result.}
By replacing the known threshold $m$ in Lemma~\ref{lem:sc_m} with the value of $m$, 
as well as setting the failure probability of exploration and satisfying Condition~\ref{cond:m} to be both $\delta/2$, we can derive to the theorem.



\section{Supplemental Experiment Results}
\label{app:exp}

\subsection{CasinoLand and RiverSwim}
\label{app:casinoland}
\begin{figure}[!htbp]
    \centering
    \includegraphics[width=0.8\textwidth]{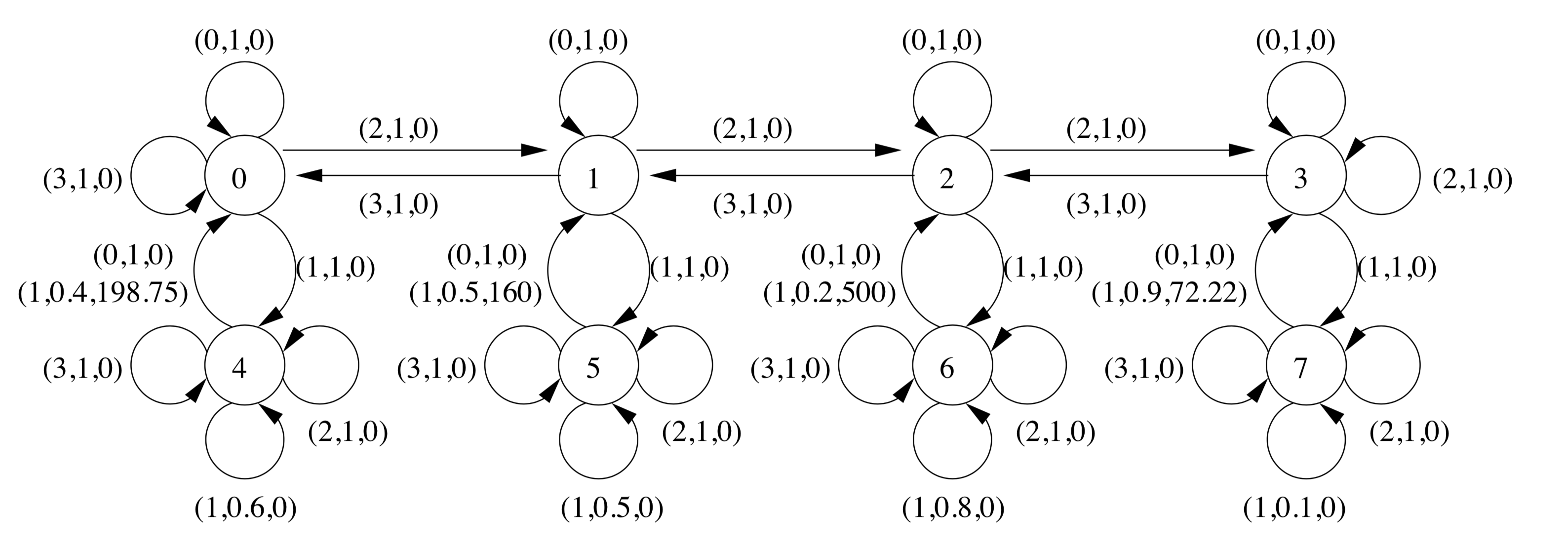}
    \caption{\textbf{CasinoLand \cite{strehl2004exploration}}}
    \label{fig:casinoland}
\end{figure}

\begin{figure}[!htbp]
    \centering
    \includegraphics[width=0.8\textwidth]{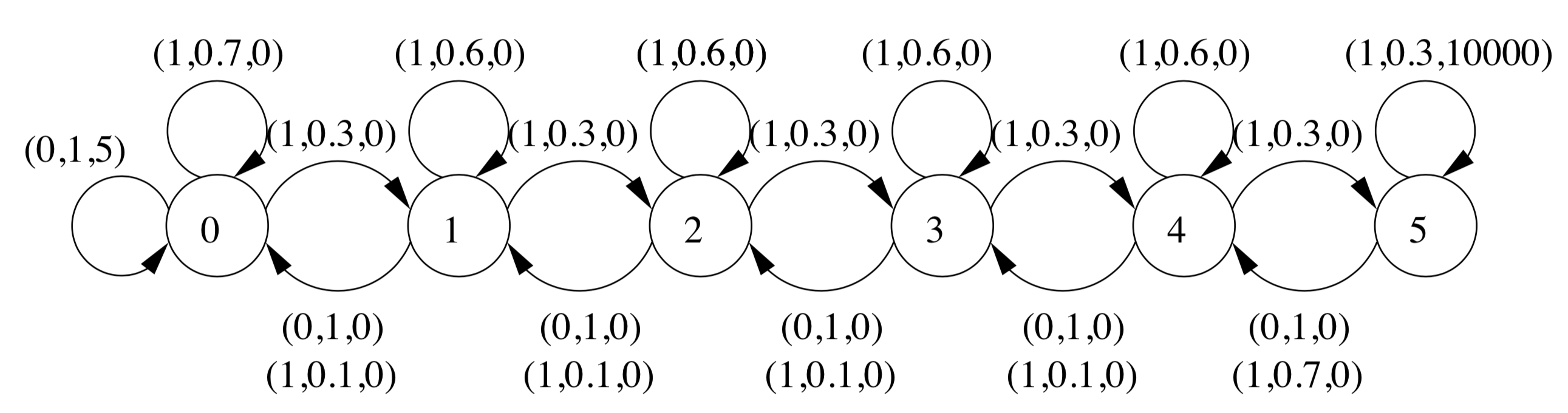}
    \caption{\textbf{RiverSwim \cite{strehl2004exploration}}}
    \label{fig:riverswim}
\end{figure}

Figure~\ref{fig:casinoland} and Figure~\ref{fig:riverswim} are the MDPs for CasinoLand and RiverSwim, which are both given by \cite{strehl2004exploration}. Note that for task CasinoLand, we assign a -100 reward for action 2 in state 4, 5, 6, and 7 to let the reward-episode lines spread out in Figure~\ref{fig:all_acc}c. 

\subsection{Selection of Known Thresholds}
\label{app:exp_th}
We design and conduct a series of systematic tests to study how the value of hyper parameters and how the MDP properties influence the learning effectiveness and efficiency of \ourmod. 
Similarly, every experiment is executed for 20 runs. The following three measurements are evaluated: (1)  \textbf{\avgrew} Average reward per episode; (2)  \textbf{\numepi} Number of episodes needed to know all states and actions; and (3) \textbf{\pavgrew} Average reward after knowing all states and actions (after exploration), which reflects accurateness of the learned dynamics.

For both RMax and \ourmod, the known threshold $m$ is the key to the practical performance of the algorithm. Although the theory tells what the optimal order of the known threshold should be, there is an unknown constant that we need to determine in practice. If the known threshold is too low, the estimated dynamics are likely to be far away from the true dynamics. But if the known threshold is set to be very high, the agent may spend too much time exploring the environment without ``knowing'' anything, which leads to low total reward in finite-episode learning.

\begin{figure}[!htbp]
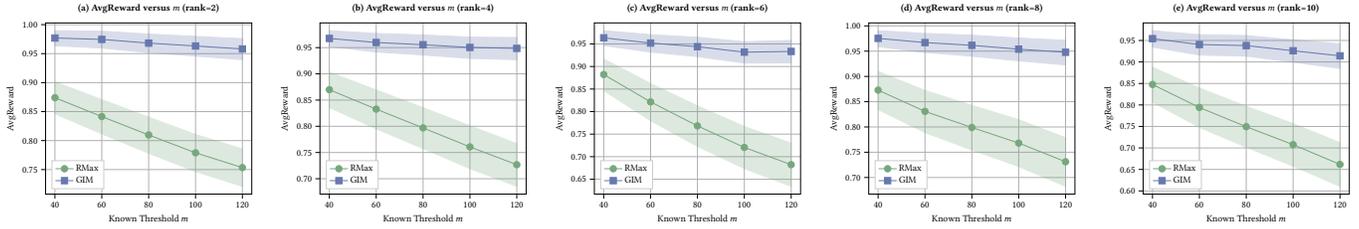

\centering
	\begin{subfigure}[t]{0.18\columnwidth}
		\centering
		\input{\fighome/appendix/avg/lowrank_S=20_A=10_r=2_runs=20.tex}	  	
	  	\label{sfig:com_avg_rank2}
	\end{subfigure}
	\hfill
	\begin{subfigure}[t]{0.18\columnwidth}
		\centering
		\input{\fighome/appendix/avg/lowrank_S=20_A=10_r=4_runs=20.tex}	  	
	  	\label{sfig:com_avg_rank4}
	\end{subfigure}
	\hfill
	\begin{subfigure}[t]{0.18\columnwidth}
		\centering
		\input{\fighome/appendix/avg/lowrank_S=20_A=10_r=6_runs=20.tex}	  	
	  	\label{sfig:com_avg_rank6}
	\end{subfigure}
	\hfill
	\begin{subfigure}[t]{0.18\columnwidth}
		\centering
		\input{\fighome/appendix/avg/lowrank_S=20_A=10_r=8_runs=20.tex}	  	
	  	\label{sfig:com_avg_rank8}
	\end{subfigure}
	\hfill
	\begin{subfigure}[t]{0.18\columnwidth}
		\centering
		\input{\fighome/appendix/avg/lowrank_S=20_A=10_r=10_runs=20.tex}	  	
	  	\label{sfig:com_avg_rank10}
	\end{subfigure}
  	\caption{Comparison of \avgrew on synthetic tasks where S=20, A=10 and rank varies from 2 to 10.} 
	\label{fig:th_compare_avg}
\end{figure}

\begin{figure}[!htbp]
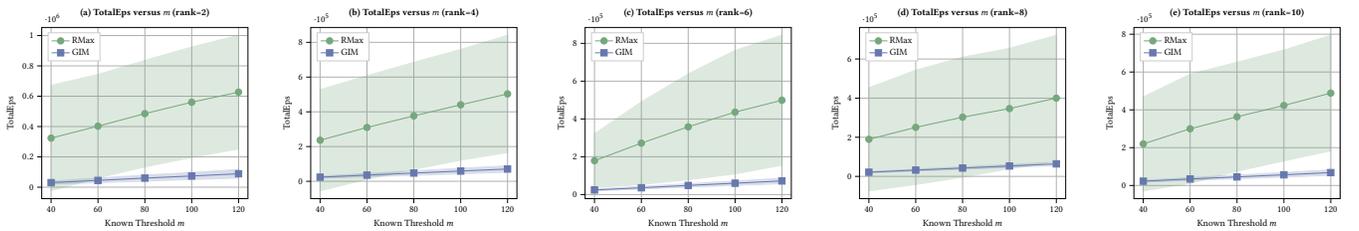

\centering
	\begin{subfigure}[t]{0.18\columnwidth}
		\centering
		\input{\fighome/appendix/times/lowrank_S=20_A=10_r=2_runs=20.tex}	  	
	  	\label{sfig:com_times_rank2}
	\end{subfigure}
	\hfill
	\begin{subfigure}[t]{0.18\columnwidth}
		\centering
		\input{\fighome/appendix/times/lowrank_S=20_A=10_r=4_runs=20.tex}	  	
	  	\label{sfig:com_times_rank4}
	\end{subfigure}
	\hfill
	\begin{subfigure}[t]{0.18\columnwidth}
		\centering
		\input{\fighome/appendix/times/lowrank_S=20_A=10_r=6_runs=20.tex}	  	
	  	\label{sfig:com_times_rank6}
	\end{subfigure}
	\hfill
	\begin{subfigure}[t]{0.18\columnwidth}
		\centering
		\input{\fighome/appendix/times/lowrank_S=20_A=10_r=8_runs=20.tex}	  	
	  	\label{sfig:com_times_rank8}
	\end{subfigure}
	\hfill
	\begin{subfigure}[t]{0.18\columnwidth}
		\centering
		\input{\fighome/appendix/times/lowrank_S=20_A=10_r=10_runs=20.tex}	  	
	  	\label{sfig:com_times_rank10}
	\end{subfigure}

  	\caption{Comparison of \numepi on synthetic tasks where S=20, A=10 and rank varies from 2 to 10.} 
	\label{fig:th_compare_times}
\end{figure}

\begin{figure}[!htbp]
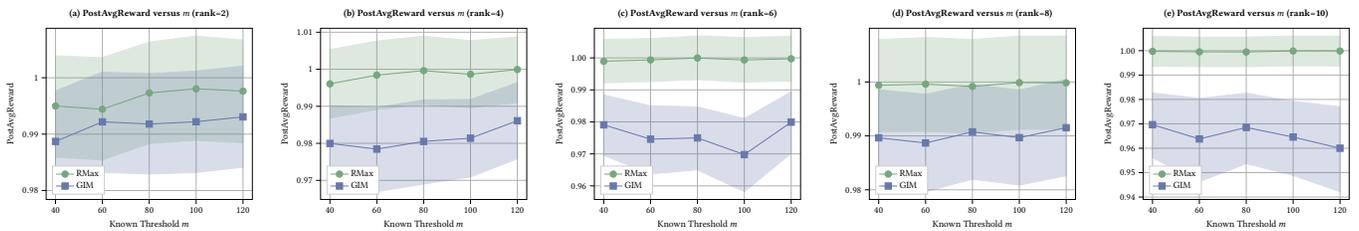

\centering
	\begin{subfigure}[t]{0.18\columnwidth}
		\centering
		\input{\fighome/appendix/pavg/lowrank_S=20_A=10_r=2_runs=20.tex}	  	
	  	\label{sfig:com_pavg_rank2}
	\end{subfigure}
	\hfill
	\begin{subfigure}[t]{0.18\columnwidth}
		\centering
		\input{\fighome/appendix/pavg/lowrank_S=20_A=10_r=4_runs=20.tex}	  	
	  	\label{sfig:com_pavg_rank4}
	\end{subfigure}
	\hfill
	\begin{subfigure}[t]{0.18\columnwidth}
		\centering
		\input{\fighome/appendix/pavg/lowrank_S=20_A=10_r=6_runs=20.tex}	  	
	  	\label{sfig:com_pavg_rank6}
	\end{subfigure}
	\hfill
	\begin{subfigure}[t]{0.18\columnwidth}
		\centering
		\input{\fighome/appendix/pavg/lowrank_S=20_A=10_r=8_runs=20.tex}	  	
	  	\label{sfig:com_pavg_rank8}
	\end{subfigure}
	\hfill
	\begin{subfigure}[t]{0.18\columnwidth}
		\centering
		\input{\fighome/appendix/pavg/lowrank_S=20_A=10_r=10_runs=20.tex}	  	
	  	\label{sfig:com_pavg_rank10}
	\end{subfigure}

  	\caption{Comparison of \pavgrew on synthetic tasks where S=20, A=10 and rank varies from 2 to 10.} 
	\label{fig:th_compare_pavg}
\end{figure}

Figure~\ref{fig:th_compare_avg}, \ref{fig:th_compare_times} and \ref{fig:th_compare_pavg} shows the \avgrew, \numepi and \pavgrew of RMax and \ourmod with different known thresholds on the synthetic tasks. To make the comparison easier to interpret, we set the average rewards obtained by the optimal policies as 1.0, and normalize the rewards of RMax and \ourmod. To make the results more convincing, we generate synthetic \dmnames with rank (roughly) equals to 2, 4, 6, 8, 10 for MDPs with 20 states and 10 actions.

In Figure~\ref{fig:th_compare_avg}, we find that within the same number of episodes, higher known thresholds result in lower per-episode reward, because more episodes are spent on exploring. However, the average reward of \ourmod is decreasing much slower than RMax, and \ourmod always keeps a very high reward rate ($>95\%$) as known threshold increases. 

The reason for their different \avgrew changes are explained by Figure~\ref{fig:th_compare_times} and \ref{fig:th_compare_pavg}. Figure~\ref{fig:th_compare_times} shows that \ourmod is able to finish exploration in much fewer episodes than RMax, and does not rise as drastically as RMax with the increasing known threshold. The fast exploration of \ourmod is not surprising because of \walkname and the completion method. But the accuracy of the dynamics returned by matrix completion remains questionable. Without effictive exploitation, efficient exploration is useless. So we also compare the corresponding \pavgrew in Figure~\ref{fig:th_compare_pavg}, where we can see, because matrix completion introduces new noises, \ourmod is not as accurate as RMax with the same known threshold. But the fact is, the difference is very small ($<1\%$). Therefore, unless the number of episodes is extremely large or infinite so that the exploration steps can be ignored, \ourmod tends to gain more rewards in total.


According to the above results, \ourmod outperforms RMax since it can get higher per-episode reward, and can find the near-optimal policy in less episodes. Futherthemore, \ourmod is less sensitive to the setting of the know threshold than RMax, while RMax needs some efforts to tune the know threshold well. Although the performance of \ourmod drops with the increasing rank, the degree of the drop is slight.

Experiments on the grid world task also demonstrate the similar results, which are all shown in Figure~\ref{fig:th_compare_grid}. 

\begin{figure}[!ht]
\centering
	\begin{subfigure}[t]{0.15\columnwidth}
		\centering
		\input{\fighome/avg/gridworld_h=4_w=4_runs=10.tex}	  	
	\end{subfigure}
	\hfill
	\begin{subfigure}[t]{0.15\columnwidth}
		\centering
		\input{\fighome/avg/gridworld_h=3_w=3_runs=10.tex}	  	
	\end{subfigure}
	\hfill
	\begin{subfigure}[t]{0.15\columnwidth}
		\centering
		\input{\fighome/times/gridworld_h=4_w=4_runs=10.tex}	  	
	\end{subfigure}
	\hfill
	\begin{subfigure}[t]{0.15\columnwidth}
		\centering
		\input{\fighome/times/gridworld_h=3_w=3_runs=10.tex}	  	
	\end{subfigure}
	\hfill
	\begin{subfigure}[t]{0.15\columnwidth}
		\centering
		\input{\fighome/pavg/gridworld_h=4_w=4_runs=10.tex}	  	
	\end{subfigure}
	\hfill
	\begin{subfigure}[t]{0.15\columnwidth}
		\centering
		\input{\fighome/pavg/gridworld_h=3_w=3_runs=10.tex}	  	
	\end{subfigure}
  	\caption{Comparison of \avgrew, \numepi and \pavgrew on $4\times 4$ and $3\times 3$ GridWorld tasks} 
	\label{fig:th_compare_grid}
\end{figure}

Based on the comprehensive comparison with RMax about known thresholds, we conclude that (1) \ourmod can explore the unknowns much faster than RMax, without sacrificing too much accuracy of predicting. (2) \ourmod is more robust to the selection of known threshold than RMax, which makes it easier to be used and tuned in practice.

\subsection{Properties of Dynamic Matrices}
\label{app:property}

\begin{figure}[!ht]
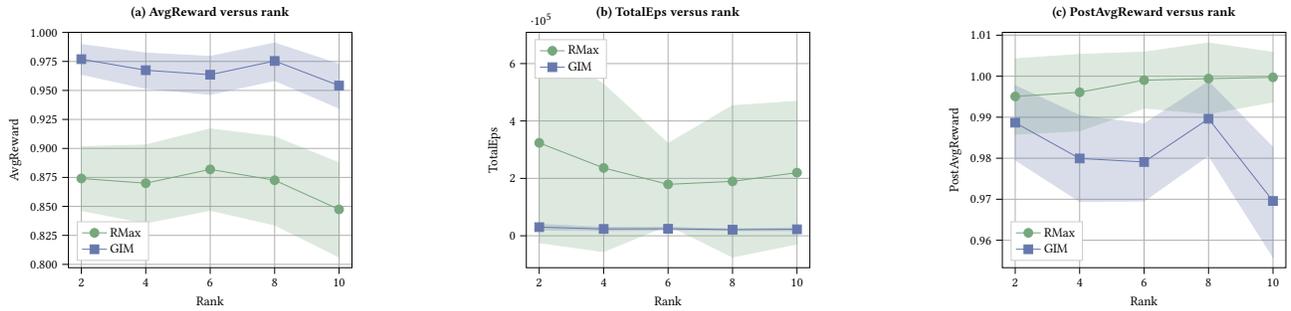

\centering
	\begin{subfigure}[t]{0.3\columnwidth}
		\centering
		\input{\fighome/appendix/rank/ranks_rew_shade_runs=20.tex}	  
	\end{subfigure}
	\hfill
	\begin{subfigure}[t]{0.3\columnwidth}
		\centering
		\input{\fighome/appendix/rank/ranks_times_shade_runs=20.tex}	  
	\end{subfigure}
	\hfill
	\begin{subfigure}[t]{0.3\columnwidth}
		\centering
		\input{\fighome/appendix/rank/ranks_post_shade_runs=20.tex}	  
	\end{subfigure}
  	\caption{Comparison of \avgrew, \numepi and \pavgrew of \ourmod and RMax on synthetic task where $S=20,A=10$ with different ranks.}
	\label{fig:ranks_full}
\end{figure}

\subsubsection{Influence of Matrix Rank $r$}

In the above section, we can see that the larger the rank is, the less accurate the return policy is (according to \pavgrew).
To make the relation between rank and the learning results more clear, we visualize how \ourmod and RMax fluctuate when the rank of the underlying \dmnames changes in Figure~\ref{fig:ranks_full}.
As the rank increases, the \numepi of \ourmod keeps stable, whereas \pavgrew drops because matrix completion works worse for high ranks. 
So \ourmod is more preferable for tasks with low-rank structures, i.e., similar transitions exist. 
But for tasks which do not have such structures, \ourmod is still able to explore efficiently and get near-optimal rewards.

\subsubsection{Influence of Incoherence Parameters $\mu_0$, $\mu_1$ and $\kappa$}

\begin{figure*}[!h]
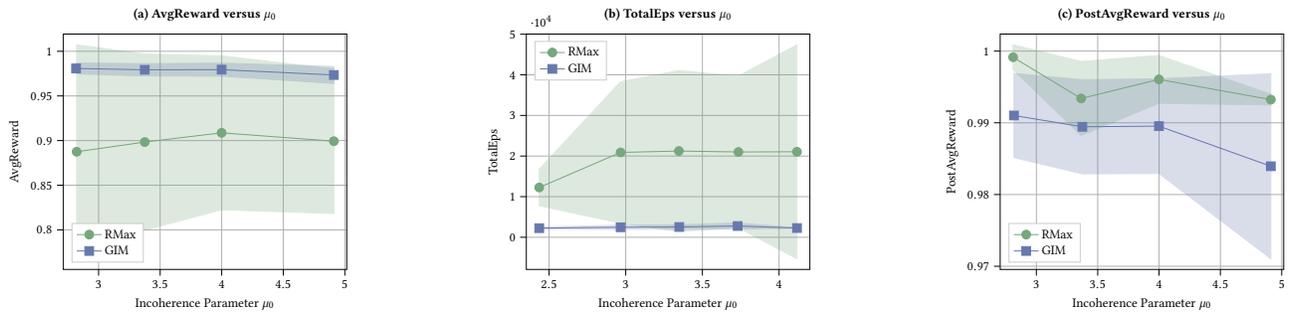

\centering
	\begin{subfigure}[t]{0.3\textwidth}
		\centering
		\input{\fighome/appendix/mus/admu_rew_line_th40_mu0.tex}  	
	\end{subfigure}
	\hfill
	\begin{subfigure}[t]{0.3\textwidth}
		\centering
		\input{\fighome/appendix/mus/admu_times_mu0_line.tex}  	
	\end{subfigure}
	\hfill
	\begin{subfigure}[t]{0.3\textwidth}
		\centering
		\input{\fighome/appendix/mus/admu_post_line_th40_mu0.tex}  	
	\end{subfigure}
  	\caption{Comparison of \avgrew, \pavgrew and \numepi of \ourmod and RMax on synthetic task where $S=20, A=10$ with different incoherence parameter $\mu_0$'s} 
	\label{fig:mu0}
\end{figure*}

\begin{figure*}[!h]
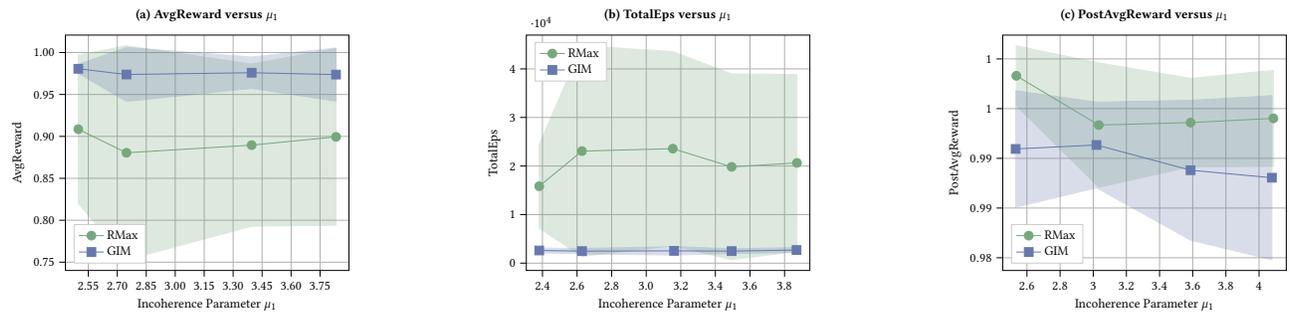

\centering
	\begin{subfigure}[t]{0.3\textwidth}
		\centering
		\input{\fighome/appendix/mus/admu_rew_line_th40_mu1.tex}  	
	\end{subfigure}
	\hfill
	\begin{subfigure}[t]{0.3\textwidth}
		\centering
		\input{\fighome/appendix/mus/admu_times_mu1_line.tex}  	
	\end{subfigure}
	\hfill
	\begin{subfigure}[t]{0.3\textwidth}
		\centering
		\input{\fighome/appendix/mus/admu_post_line_th40_mu1.tex}  	
	\end{subfigure}
  	\caption{Comparison of \avgrew, \pavgrew and \numepi of \ourmod and RMax on synthetic task where $S=20, A=10$ with different incoherence parameter $\mu_1$'s} 
	\label{fig:mu1}
\end{figure*}

\begin{figure*}[!h]
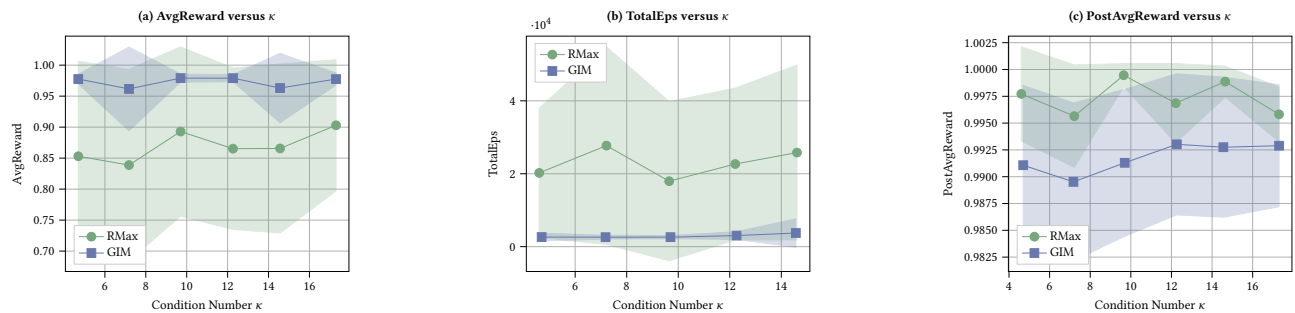

\centering
	\begin{subfigure}[t]{0.3\textwidth}
		\centering
		\input{\fighome/appendix/mus/admu_rew_line_th40_kappa.tex}  	
	\end{subfigure}
	\hfill
	\begin{subfigure}[t]{0.3\textwidth}
		\centering
		\input{\fighome/appendix/mus/admu_times_kappa_line.tex}  	
	\end{subfigure}
	\hfill
	\begin{subfigure}[t]{0.3\textwidth}
		\centering
		\input{\fighome/appendix/mus/admu_post_line_th40_kappa.tex}  	
	\end{subfigure}
  	\caption{Comparison of \avgrew, \pavgrew and \numepi of \ourmod and RMax on synthetic task where $S=20, A=10$ with different condition number $\kappa$'s} 
	\label{fig:kappa}
\end{figure*}

Figure~\ref{fig:mu0}, Figure~\ref{fig:mu1} and Figure~\ref{fig:kappa} respectively represent the \avgrew, \numepi and \pavgrew of \ourmod and RMax on a series of synthetic tasks with different incoherence parameters ($\mu_0, \mu_1$) and condition number $\kappa$. Based on the figures, we conclude that \ourmod does not change a lot with $\mu_0, \mu_1$ and $\kappa$, in terms of efficiency and accuracy. 

\end{document}